\documentclass[journal]{IEEEtran}

\usepackage{amssymb}
\usepackage{amsmath,mathtools}
\usepackage{hyperref}
\usepackage{graphicx}
\usepackage[table,xcdraw]{xcolor}
\usepackage{booktabs}
\usepackage{algorithm}
\usepackage{algpseudocode}
\usepackage{caption}
\usepackage{lipsum}
\usepackage{multirow}
\usepackage{relsize}
\usepackage{algorithm}
\usepackage{algpseudocode}
\usepackage{tabularx}
\usepackage{cleveref}
\usepackage{bm}

\usepackage{amsfonts}
\usepackage{array}
\usepackage[caption=false,font=normalsize,labelfont=sf,textfont=sf]{subfig}
\usepackage{textcomp}
\usepackage{stfloats}
\usepackage{url}
\usepackage{verbatim}
\def\BibTeX{{\rm B\kern-.05em{\sc i\kern-.025em b}\kern-.08em
    T\kern-.1667em\lower.7ex\hbox{E}\kern-.125emX}}
\usepackage{balance}
\usepackage{todonotes}

\DeclareMathOperator*{\argmax}{arg\,max}

\newcommand\changes[1]{\textcolor{black}{#1}}
\newcommand\vect[1]{\boldsymbol{#1}}

\begin{document}


\title{Deep Reinforcement Multi-agent Learning framework for Information Gathering with Local Gaussian Processes for Water Monitoring}

  \author{Samuel~Yanes~Luis\IEEEauthorrefmark{1},
      Dmitriy~Shutin\IEEEauthorrefmark{2},
      Juan~Marchal~Gómez\IEEEauthorrefmark{2},
      Daniel~Gutiérrez~Reina\IEEEauthorrefmark{1},
      and~Sergio~Toral~Marín\IEEEauthorrefmark{1}
    \\
    \IEEEauthorblockA{\IEEEauthorrefmark{1} Department of Electronic Engineering, University of Sevilla}\\
    \IEEEauthorblockA{\IEEEauthorrefmark{2} Institute of Navigation and Communications,  German Aerospace Center}%
  }  


            
            




\markboth{arXiv}{Yanes \MakeLowercase{\textit{et al.}}: Deep Reinforcement Multi-agent Learning framework for Information Gathering with Local Gaussian Processes for Water Monitoring}

\maketitle


\begin{abstract}

\IEEEPARstart{T}{he}  conservation of hydrological resources involves continuously monitoring their contamination. A multi-agent system composed of autonomous surface vehicles is proposed in this paper to efficiently monitor the water quality. To achieve a safe control of the fleet, the fleet policy should be able to act based on measurements and to the the fleet state. It is proposed to use Local Gaussian Processes and Deep Reinforcement Learning to jointly obtain effective monitoring policies. Local Gaussian processes, unlike classical global Gaussian processes, can accurately model the information in a dissimilar spatial correlation which captures more accurately the water quality information. A Deep convolutional policy is proposed, that bases the decisions on the observation on the mean and variance of this model, by means of an information gain reward. Using a Double Deep Q-Learning algorithm, agents are trained to minimize the estimation error in a safe manner thanks to a Consensus-based heuristic. Simulation results indicate an improvement of up to 24\% in terms of the mean absolute error with the proposed  models. Also, training results with 1-3 agents indicate that our proposed approach returns 20\% and 24\% smaller average estimation errors for, respectively, monitoring water quality variables and monitoring algae blooms, as compared to state-of-the-art approaches.

\end{abstract}

\IEEEpeerreviewmaketitle

\section{Introduction}

Environmental monitoring is vital for assessing various aspects of the natural world, including air, water, and soil quality, biodiversity, and climate. This practice plays a crucial role in addressing environmental issues such as pollution, climate change, and biodiversity loss \cite{alvarez-rogel_case_2020}. By focusing on key indicators like air and water quality, scientists and policymakers can pinpoint pollution sources and implement effective measures to reduce or eliminate them, informing the development of impactful environmental policies and regulations. To efficiently monitor water resources, autonomous surface vehicles (ASVs) emerge as a promising solution, particularly for vast geographic areas challenging to survey manually \cite{ASVsReef}. ASVs offer a cost-effective means of data collection using diverse sensors, including physicochemical sensors and cameras \cite{YuanReview, Hu_2021}. Leveraging these autonomous agents allows for the swift acquisition of comprehensive environmental data, enhancing our understanding of environmental conditions accurately and efficiently.

This article focuses on a specific issue in natural conservation, building upon prior research \cite{kathen2022aquafelpso, peralta_MULTIOBJECTIVE}. It addresses the monitoring of Ypacaraí Lake, Paraguay's largest drinking water source, using autonomous surface vehicles (ASVs). Ypacaraí's water quality, crucial for the population, is characterized by variables like pH, turbidity, dissolved oxygen, and chlorophyll concentration. Contamination sources vary due to geographical, human, and biological factors, influencing the optimal information acquisition process. Monitoring the contamination is challenging due to their localized and temporal occurrence, influenced by wind and tides within the lake. This article proposes the adaptive deployment of a multi-agent vehicle fleet capable of efficiently monitoring common water quality phenomena such as algae blooms.

\changes{The approach proposed in this paper for monitorization of biological phenomena will fall into the category of the so-called Informative Path Planning (IPP) \cite{popovic_informative_2020}. The ultimate goal of this family of problems is to sequentially decide on an optimal path for every vehicle that maximizes the information $\mathcal{I}$ collected, while minimizing the cost $\mathcal{C}$ associated with data acquisition. However, when agents have little or no information about the environment, it is difficult to make effective decisions in the presence of such high uncertainty.} This problem has been previously addressed in works such as \cite{peralta_MULTIOBJECTIVE}, where by means of Bayesian optimization (BO) and an Acquisition Function, the next position for data acquisition is chosen considering the overall uncertainty and the current information model.

IPP strategies are intrinsically adaptive as, depending on the successive observations of the scenario, the agents must change their monitoring waypoints. When different cooperating agents are considered for monitoring, other characteristics come into play that increase the complexity of the IPP. This is the case of redundancy in measurement or collision avoidance between agents. In the first case, it is obvious that when paths are constrained to a distance and a time budget, taking redundant measurements is against efficiency. Agents must coordinate by means of some method to avoid oversampling areas of which they already have sufficient knowledge or to simultaneously monitor the same area. To address these multiple aspects of IPP, \changes{this paper proposes} to embed this IPP into a Reinforcement Learning framework \cite{sutton_reinforcement_2018}, that explicitly considers the information gain with the redundancy measure of every vehicle. The other important aspect in multiagent IPP is agent-agent and agent-environment collision avoidance. which is an aspect usually neglected in previous works  \cite{kathen2022aquafelpso}. 

Second, the GPs provide a probabilistic description of the process in terms of its mean $\vect{\mu}$ and covariance $\vect{\sigma}$, which naturally provides a measure of uncertainty about the learned process.


However, GPs have certain issues that make them difficult to work with. First, they have a complexity of $O(N^3)$, with $N$ being the number of measurement samples \cite{RasmussenBible}. Moreover, the use of GPs in environmental monitoring typically assumes constant hyperparameters across the entire data range.In the case of most kernels used in the literature \cite{peralta_MULTIOBJECTIVE}, the length scale hyperparameter defines the size of spatial correlation between samples of the process. 
Having the same length scale implies constant spatial correlation properties across the whole area of interest. However, practically, this is often not the case, as e.g. concentrations of the algae can change quite abruptly. 


Therefore, another major contribution of this work is the use of multiple local GPs. Local GPs have only local influence, so they will fit the data seen under their area of influence. It will be shown that local GPs are able to characterize better areas with distinct length scales. Furthermore, local GPs bring a considerable improvement in the scalability of the algorithm, since the complexity is reduced in global terms from $O(N^3)$ to $O(N^3 / M ^ 2)$, with $M$ being the number of local GPs. 


Finally, to solve the IPP, the use of Deep Reinforcement Learning (DRL) techniques to train deep adaptive policies. In recent years, DRL has begun to be used for multiagent \cite{yanes_multiagent_2021} path planning. DRL allows a neural network to optimize a multi-agent policy to maximize a long-term objective set in a reward function. The reward function acts as a measurement of the optimality of each action $a$ given an observation of the environment $o$. Through the interaction of agents and the environment, DRL algorithms such as Deep Q-Learning (DQL) \cite{mnih_human-level_2015} are able to adapt the fleet behavior of vehicles to improve acquisition in an optimization time horizon. For this application, the use of DRL is also convenient because it will allow behavior specialization of a fleet for a broad set of scalar fields by means of realistic simulators of the environments within the known boundaries of each biological process.

Finally, \changes{the framework is validated} against other path planning algorithms in the different benchmarks. 


In summary, the novelties proposed in this article are 

\begin{itemize}
    \item A Local Gaussian Process model for multimodal environmental scenarios.
    \item A DRL framework to maximize the gathered information for a  fleet of unmanned vehicles, including the reward function based on the information gain and the observation method.
    \item A censoring methodology to avoid agent-agent, agent-environment collisions.
\end{itemize}

\changes{This paper is organized as follows: In Section II, previous approaches and advances in Information Gathering and path planning with autonomous vehicles are discussed. In Section III, the problem of Informative Path Planning is described and the Ground Truth under monitorization as well. Later, in Section IV, the methodology is explained. First, the Local Gaussian Process proposition is explained, and later, the Deep Reinforcement Learning algorithm and formulation is described. In Section V, the different results and simulation are described, and also the comparison with other path planning techniques. Finally, in Section VI, the conclusions are presented with future lines of works that shall be addressed.}

\section{Related Work}


The use of autonomous aquatic and aerial vehicles has gained relevance in recent years due to advances in battery autonomy and, above all, to the capability for remote computing and sensing \cite{sanchez-garcia_survey_2018}. In the particular case of Aquatic autonomous vehicles, they are divided into two types: i) surface vehicles (ASV) and ii) underwater vehicles (USV). The former has been used especially for monitoring water quality parameters (WQP) in rivers, lakes, and coasts. 
The use of autonomous surface vehicles for the acquisition of information in natural environments has been gaining momentum lately. In several previous works, \changes{it is possible to find examples} multiagent \cite{peralta_MULTIOBJECTIVE} applications for this purpose. In general, these vehicles are equipped with sensor modules for the acquisition of WQP and can take geographically located samples at one point at a time.

Another common approach is the use of algorithms based on Particle Swarm Optimization (PSO) \cite{kathen_informative_2021}. PSO algorithms base agent decisions on swarm behavior. In \cite{kathen_informative_2021}, the use of GPs to improve the classical PSO algorithm is also discussed. In this approach, vehicles update their speeds attracted by the points of highest uncertainty of the GP, the highest individual's, collective and estimated pollution values. This approach results in explorative-exploitative paths and a very computationally scalable algorithm. 
\changes{This approach is not always adequate due to local gradient continuity fluctuations, especially for the case of algae bloom monitoring where the information gradient is discontinuous.} 


In relation to the use of deep reinforcement learning for path planning, is easy to find an upward trend in the number of recent articles such as \cite{yanes_multiagent_2021}. Previous approaches such as \cite{yanes_deep_2020} have focused on solving the informative patrol problem for Ypacaraí Lake itself. This problem consists of continuous monitoring of WQP with a temporal cyclic criterion. In \cite{yanes_multiagent_2021} a multiagent version of the Double Deep Q-Learning (DDQL) algorithm of \cite{mnih_human-level_2015} is used. The DDQN algorithm uses two equal neural networks to estimate the cumulative future reward given a state observation $o$ and the set of possible actions $a$. From this work, it is possible to see that the neural decoder structure with visual observations is used, similar to our proposal. This visual formulation of the state allows for better interpretability and simplifies the feature selection process for the estimation of the estimated future reward. However, unlike our new proposal, the state is completely known a priori, which is unrealistic in an initial exploration scenario such as the one that in this paper's proposal. 


In \cite{viseras_deepig_2019} a work based on DRL is also presented to solve IPP with multiple agents. The objective is to reduce the estimation error over a relatively small scalar field ($10 \times 10$ pixels) in the minimum possible time. Similar aspects between this approach and ours are: i) the use of partial and visual observations of the environment, and ii) the use of discrete actions as a way of reducing the decision variables of each agent. Whereas they propose a Dueling Deep Q-Learning, this work proposes the Proximal Policy Optimization (PPO) algorithm which uses rollouts of experiences of the neural policy $\pi(\vect{s})$ to update its parameters. A reward function based on the Root Mean Squared Error (RMSE) is proposed, which implies that the reward can only be calculated in simulation, when the ground truth is known. Our proposal attempts to decouple the reward from an error function, which is associated with an unknown ground truth, and the error function itself. 


Other approaches using DRL for path planning have focused solely on obstacle avoidance. In \cite{zhang_autonomous_2022} one can find a promising example of the use of DRL techniques such as Advantage Actors Critic (A2C) for the generation of obstacle-free paths with a single flying drone. In this paper, obstacles can be both dynamic and static. The policy learns to avoid obstacles by internalizing the obstacles. Another example of obstacle avoidance mechanism is found in \cite{censoring_yanes_2022}, where DDQL is used to solve the IPP with a single agent.


\section{Statement of the problem}


IPP is defined as a sequential decision process in which the objective is to maximize the information $I(t)$ collected over time. For the multi-agent case, a set of paths $\Psi:=  \left[ \psi_1, \psi_2, ..., \psi_N  \right]$ that maximize the joint information will be sought with the restriction that these paths have no agent-agent or agent-obstacle collisions:

\begin{equation}
    \Psi^* = \argmax_{\Psi} \sum_{\psi \in \Psi} I(t)_{\psi}
\end{equation}

In our approach, a path will be defined by a series of measurement points $X_j^{meas}$ for each agent $j$ in a fleet of $N$ agents. Each vehicle will take at each instant $t$ an action $\vect{a}$ from the possible set of actions $A$. These actions correspond to the 8 vectors of movement $[S,SE,E,NE,N,NW,W,SW]$. Each action involves moving the agent in that direction over a fixed distance $d_{meas}$, and taking a water sample wherever the vehicle is. The paths will have a maximum length of $d_{max}$ until the battery level reaches a safety threshold that prevents them from returning to the base. Samples are taken from a ground truth that is a static scalar field $Y$ predefined at the start of the mission but unknown except at those points where the agents sample. 
The measurement model for vehicles is represented as 
\[
    y_p = Y(\vect{p}) + \epsilon_p,
\]
where $\vect{p} = \left[ p_{lat}, p_{long}\right]$ is a vehicle position, $\epsilon_p$ is a noise associated with the variability of sensor measurement, and $Y: \mathbb{R}^2 \rightarrow \mathbb{R}$ is the sought-after (ground truth) function. 
It will be assumed that no vehicle can take samples or visit areas with obstacles. 
The navigation limits of the ASVs are indicated by a navigation map $M: \mathbb{R}^2 \rightarrow \mathbb{R}$, where $M(p) = 0$ in the areas that are unreachable. Additionally, no vehicle may be in the same zone at the same time. Two vehicles are considered to be in the same zone when the distance between them is less than $d_{\mathrm{safety}}$. This is a strong constraint to ensure the safety and integrity of the fleet, in addition to reducing redundancy in the measurement. 
In relation to the initial point, each ASV $j$ has a valid deployment zone on each map $\mathcal{Z}^j$. No vehicle can start outside its zone due to coastal security restrictions. It will be assumed that the values of $Y$ are normalized between 0 and 1, called the Normalized Contamination Index (NCI), for better comparison between applications. In this case, a value of 0 is considered a value outside of any biological risk, and 1 a value with high biological risk or high water contamination. 

Given this set of hypotheses and statements, it is possible to formulate the problem as a Partially Observable Markov Decision Process (POMDP). 
This type of decision process can be expanded to the multiagent case by defining multiple agents and a partially observable state $s_t$, only accessible through an observation function $o_j^t = \mathcal{O}(s^t)$ by the agent $j$. 
The optimization goal of any POMDP is to find the optimal policy $\vect{a}_j^{t+1} = \pi^*(o_j^t)$ that maps an observation $o_j^t$ into an action that maximizes the accumulated reward given by $\sum_{t=0}^T R(s^t, \vect{a}_j^t)$ and over some optimization time budget $T$ for each agent $j \in [1,N]$ in the fleet. 
From the complete state $s^t$ that gathers all the information for the environment, the complete ground truth scalar field is ignored, and it is only possible to know at each instant $t$ the positions of the ASVs, the navigation map $\mathcal{M}$, and the samples that have been taken so far by each vehicle $\{Y(X^{\mathrm{meas}}), X^\mathrm{meas} \}$.

\subsection{Ground Truths models}

\begin{figure}[th]
    \centering
    \includegraphics[width=0.6\linewidth]{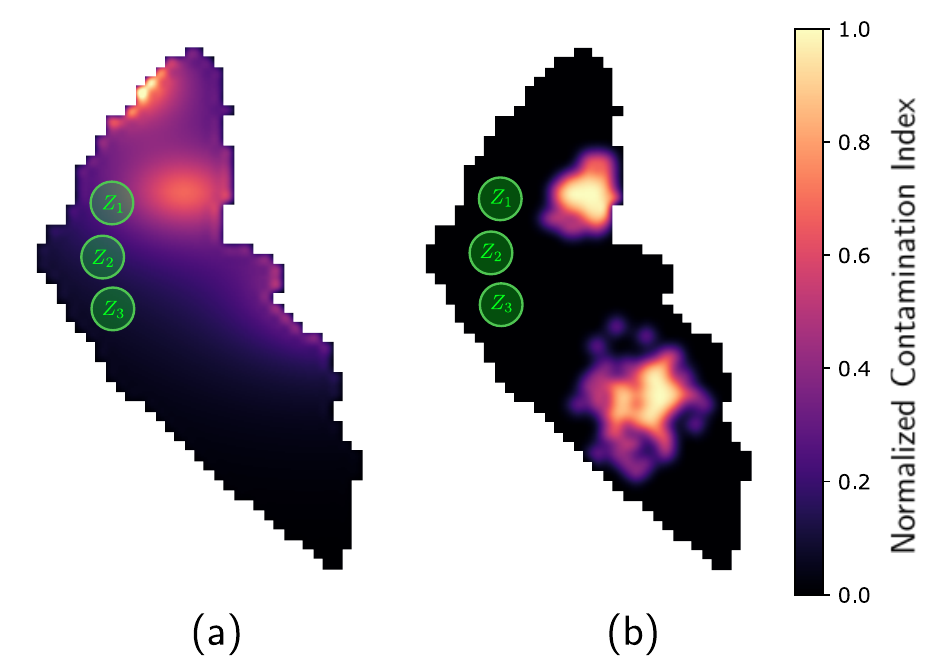}
    \caption{Example of the ground truths used for every mission. In (a), the WQP map. In (b) an example of an algae scenario with two blooms. In green, $Z_1, Z_2, Z_3$ correspond to the initial deployment zones of the vehicles. The initial position of every vehicle is randomly selected within this areas.}
    \label{fig:gts}
\end{figure}

This article focuses on two cases of contamination for natural water resources. First, it is the case of monitoring smoothly distributed physicochemical parameters (WQP) such as pH, dissolved oxygen, Oxidation-Reduction Potential (ORP), etc. As described in \cite{peralta_MULTIOBJECTIVE}, these parameters can be characterized as mountains and valleys of different heights randomly distributed on the navigable surface (see Figure \ref{fig:gts}a). 
For the second case, in the algae blooms case, the distribution of cyanobacterial clusters is more localized, exhibiting higher variation of gradients at some locations. Random blooms can occur at different points in the lake, causing several hot spots of pollution (see Figure \ref{fig:gts} b). Chlorophyll or turbidity sensors are often used to measure them. These blooms also respond to the dynamics of the tide and wind, acting like surface particles. The framework starts by simulating the dynamics of the algae blooms for a random amount of time prior to any mission to start with any possible state during the algae dissemination process. In the end, it is proposed to model both phenomena in two ground truth generators that provide the learning algorithm with randomly generated scalar functions $Y$. 
In both cases, the functions are considered static since the dynamics of the parameters to be measured is much slower than the total time to complete a mission (about 4h).

\subsection{Assumptions}

The following assumptions are used throughout this article:

\begin{itemize}
    \item The navigable waters map will be the same from one episode to another and is obtained by the real navigation map of the Water Resource under monitorization.
    \item This article assumes that vehicles must be homogeneous, with equal movement and measurement capabilities.
    \item The vehicles can take the actions and reach the target points without any problem. No moving obstacles are considered within the scenario other than the existence of the other vehicles.
    \item Vehicles do not have to end up in the same place as they started. They are considered to have sufficient autonomy to return to shore from any point at the end of the monitoring.
    \item It is necessary to have a prior approximate behavioral model of the information. 
    \item Measurement noise from vehicles is considered to be negligible.
\end{itemize}

\section{Methodology}

In this section, the methodology used to i) perform the model estimation online with local GPs and ii) train the ASVs policy using DRL, is described.

\subsection{Local Gaussian Process for estimation}
\label{section:gps}

\begin{figure}
    \centering
    \includegraphics[width=1\linewidth]{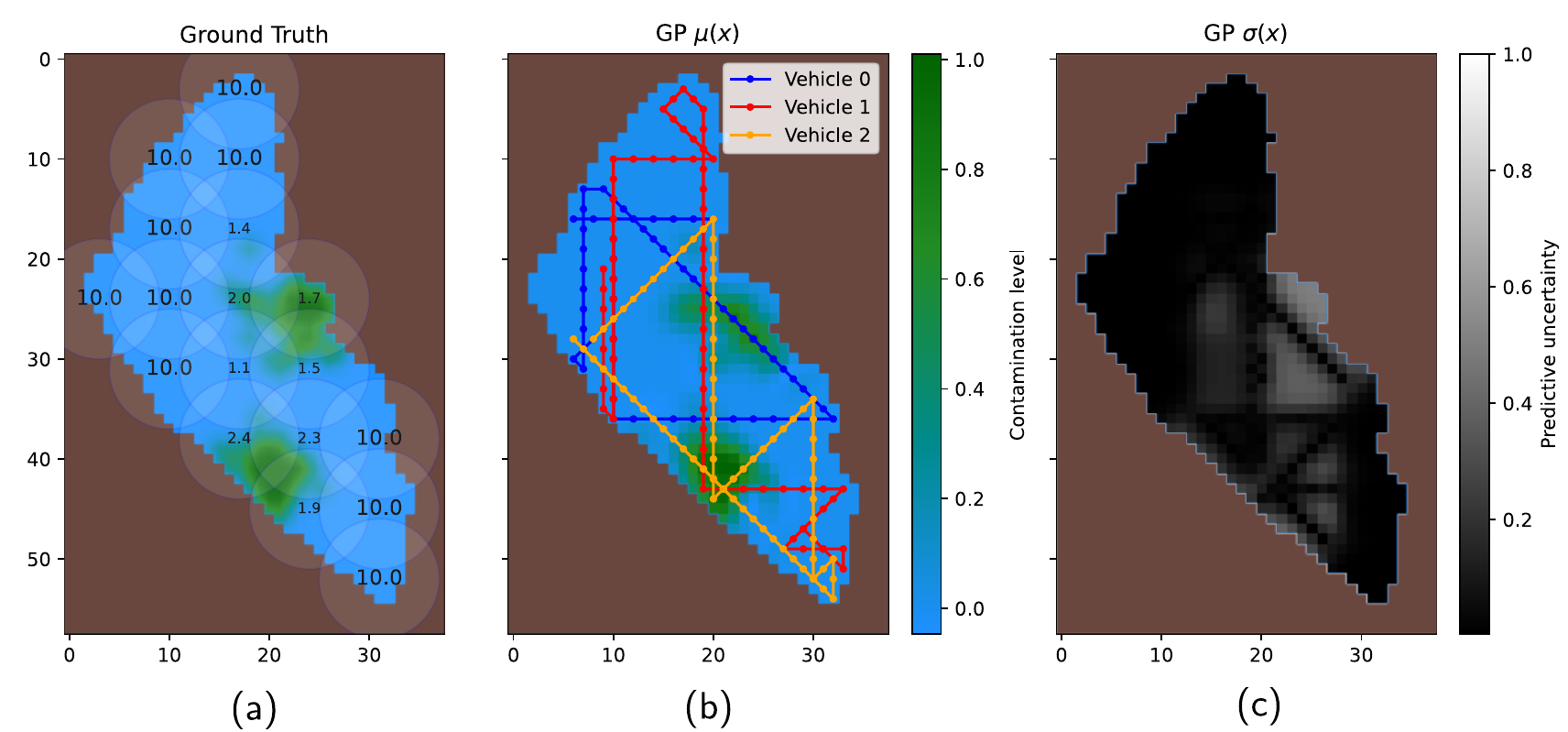}
    \caption{Local GP applied to algae bloom detection with random paths for 3 ASVs. In (a), the local GpS influence areas and the Ground Truth. In (b) the synthesized model from the local GP $\hat \mu(x)$. In (b), the joint predictive uncertainty $\hat \sigma(x)$.}
    \label{fig:Zones}
\end{figure}

A Gaussian Process (GP) is a stochastic process that is fully specified by a mean function $\mu(x)$ and a covariance function $k(x,x')$. 
GPs are used in this work, similarly to \cite{peralta_MULTIOBJECTIVE}, as online prediction methods to obtain a contamination model. 
This model will serve, on the one hand, as an estimation of WQP/algae concentration and, on the other hand, as an observer of the hidden state of the POMDP. A GP is denoted by:

\begin{equation}
    f(\vect{x}) \sim GP(\mu(\vect{x}), k(\vect{x},\vect{x}'))
\end{equation}

A Gaussian Process (GP) is a nonparametric probabilistic approach that can be used to estimate an unknown function $f(X)$, given a set of input-output pairs $(\vect{X}, \vect{y})$ \cite{RasmussenBible}. In the case of water quality parameters, $\vect{X}$ represents the sampling locations, and $\vect{y}$ represent the measured water quality parameters at these locations. A GP assumes that the joint distribution of the output values $\vect{y}$ is Gaussian, with a mean function $\mu(\vect{x})$ and a covariance function $k(\vect{x},\vect{x}')$ that specifies the similarity between any $\vect{x}$ and $\vect{x}'$ in the input space. The GP prediction for a set of measured points $\{\vect{X}^{meas}, \vect{y}\}$ is given by:

\begin{equation}
\begin{aligned}
    \hat{f}\left(X\right) & =\vect{k}_*^T\left(\vect{K}+\sigma_n^2 \vect{I}\right)^{-1} \vect{y} \\
    \text { where } \quad \vect{k}_* & =\left[k\left(\vect{x}^*, \vect{x}^{meas}_1\right), \ldots, k\left(\vect{x}^*,\vect{x}^{meas}_n\right)\right]^T \\
    \vect{K} & =\left[k\left(\vect{x}^{meas}_i, \vect{x}^{meas}_j\right)\right]_{i, j=1}^n
\end{aligned}
\end{equation}

In this equation, $\mathbf{k}^*$ represents the vector of kernel evaluations between the new input point $\vect{x}^*$ and all training input points $\vect{x}^{meas}_i$, $i=1,\ldots,n$. $\mathbf{K}$ is the covariance matrix between all training input points, and $\sigma_n^2$ is the noise variance of the observations. The GP prediction can be conditioned on observed data by incorporating the training data in the mean and covariance functions. Specifically, given the set of training data $(\vect{X}^{meas}, \mathbf{y})$, where $\vect{X} = [\mathbf{x}_1^{meas}, \ldots, \mathbf{x}_n^{meas}]^T$ and $\mathbf{y} = [y_1, \ldots, y_n]^T$, the posterior distribution of the uncertainty values at a new input point $\vect{x}*$ can be written as:

\begin{equation}
\vect{\sigma}^2(\vect{x}^*) = k\left(\vect{x}^*, \vect{x}^*\right)-\mathbf{k}_*^T\left(\mathbf{K}+\sigma_n^2 \mathbf{I}\right)^{-1} \mathbf{k}_*
\end{equation}

In this equation, $\vect{\sigma}^2(\vect{x}^*)$ represents the variance of the posterior distribution of the function value at a possible new  measurement point. 

\changes{The choice of the kernel function defines how the input variables are correlated. An immediate choice for the task of WQP monitoring, as explained in \cite{peralta_MULTIOBJECTIVE}, is to use an RBF-type kernel, which reduces the correlation exponentially with the distance between samples. This kernel, then, imposes a smooth structure modeled by its lengthscale, which in most cases it is sufficient to conform a good model}:

\begin{equation}
    k_{RBF}(\vect{x},\vect{x}') = \sigma_0^2 \exp\left({\frac{-\|\vect{x}-\vect{x}'\|^2}{2\ell}}\right)
\end{equation}

In GPs, the hyperparameters $\vect{\theta} = (\sigma_0^2, \ell)$ are learned from the training data by maximizing the type-II log-likelihood function, which is the likelihood of the hyperparameters given the observed data $\{\vect{X}^{\mathrm{meas}},\vect{y}\}$ \cite{RasmussenBible}.

However, in the classical approach where the whole search space $\vect{X}$ is estimated with the same GP, it is implicitly assumed that the same hyperparameters, e.g. $\ell$ defining the smoothness of the estimated function $\hat f(\vect{x})$, are valid for all regions of the explored environment. 
This, as will be seen later, may be incorrect for functions with local behaviors.

To deal with benchmark function with different levels of continuity and gradient smoothness, such as those in Figure \ref{fig:gts}, the use of Local GPs is proposed. Local GPs consist of a set of GPs that are only valid on a subset of the total search space $\vect{X}_{local}$. 
First, to this end it is possible to define a set of centroids $\vect{c}_k$ homogeneously distributed over the map. 
Each centroid defines a GP and has a radius of action of $\nu_k$. 
Thus, a model with $K$ local GPs is defined as:

\begin{equation}
\begin{aligned}
    GP_{local} := \{ GP_1, \ldots, GP_K\} \\
    \text{where: } \quad GP_k = GP(\vect{\mu}, k, \vect{c}_k, \nu_k)
\end{aligned}
\end{equation}

To improve consensus between local GPs when fitting the hyperparameters online, several shared zones of influence surge, where a sample $(\vect{x},y)$ is used to fit several processes simultaneously (see Figure \ref{fig:Zones}a). Shared areas guarantee better smoothness in the limits of the local areas. 
The level of redundancy and the granularity of the processes is defined with $(\nu_k, \vect{c}_k)$ parameters, and will be adjusted for each case.

To compute a total joint mean $\vect{\hat \mu}(\vect{x})$ and uncertainty $\vect{\hat \sigma}(\vect{x})$ for a point $\vect{x}$, a weighted mean among all values assigned by the GPs will be used. 
The influence of each GP on a value x, will be proportional to the distance from that location to the centroid of that GP. This is convenient since the whole model can easily reach a consensus between GPs because they often see the same data in shared zones, while being robust against outliers or a possible non-convergence of a local GP where a new sample does not improve a particular GP. This joint model, as depicted in Figures \ref{fig:Zones}a and \ref{fig:Zones}b, is transparent to the sampling process and produces an output with the same size of a global GP. Thus, the joint mean and uncertainty $(\vect{\hat \mu}(\vect{x}), \vect{\hat \sigma}(\vect{x}))$ can be defined as:

\begin{equation}
\begin{aligned}
    \vect{\hat \mu}(\vect{x}) =
                        \mathlarger{\mathlarger{\sum}_{i = 1}^{K}} \mu_i(\vect{x}) e^{-\|\vect{x}-\vect{c}_i\|_2}
                        \left(\mathlarger{\mathlarger{\sum}_{i = 1}^{K}} e^{-\|\vect{x}-\vect{c}_i\|_2}\right)^{-1}
\end{aligned}
\end{equation}

\begin{equation}
\begin{aligned}
    \vect{\hat \sigma}(\vect{x}) = 
                        \mathlarger{\mathlarger{\sum}_{i = 1}^{K}} \sigma_i(\vect{x}) e^{-||\vect{x}-\vect{c}_i||_2}
                        \left(\mathlarger{\mathlarger{\sum}_{i = 1}^{K}} e^{-||\vect{x}-\vect{c}_i||_2}\right)^{-1}
\end{aligned}
\end{equation}

\changes{Any Gaussian model also allows the imposition of a prior on the information obtained. In both Ground Truths, a prior of mean 0 is imposed, assuming that the measurements are normalized. This makes it possible to establish that, in the absence of measurements and when uncertainty is at a maximum, leading to zero correlation, the estimated value at those points coincides with the prior. The proposed local Gaussian processes follow the same rule. This implies some knowledge of the scalar field to be measured. Any other type of Ground Truth should consider how the information behaves a priori. The same applies to hyperparameters. A range of possible hyperparameter values has to be considered with which to start the optimization of each GP. In our case, the initial value of all processes is chosen to be the maximum possible $\ell_{0} = 10$, indicating that, a priori, the information is smooth. The interval of values imposed is $(\ell_{min}, \ell_{max}) = (0.1, 10)$, so that in the optimization, l will be kept bounded.}
     
\subsection{Deep Reinforcement Learning}

Deep Reinforcement Learning (DRL) is a sub-field of machine learning that combines deep neural networks with reinforcement learning to enable agents to learn to make optimal decisions in complex environments \cite{mnih_human-level_2015}. In DRL, an agent interacts with an environment and receives rewards or penalties for its actions. The goal of the agent is to learn a policy that maximizes the expected cumulative reward over time \cite{sutton_reinforcement_2018}. In this article, the algorithm Double Deep Q-Learning (DDQL) is proposed as a common and successful framework to optimize discrete action policies \cite{yanes_deep_2020, yanes_multiagent_2021, censoring_yanes_2022}. DDQL uses a deep neural network to approximate the action-value function $Q(s,\vect{a})$ \cite{mnih_human-level_2015}. The action-value function is a function that maps a state-action pair to the expected cumulative reward. The Q-learning algorithm uses an iterative process to update the Q-values based on the observed rewards and the discounted future rewards. The updates are given by the Bellman equation \cite{bellman}:


\begin{equation}
\begin{split}
    Q(s_t, \vect{a}_t) \leftarrow  \\ 
    Q(s_t, \vect{a}_t) + \alpha \left[ r_{t+1} + \gamma \max_{\vect{a}'} Q^{target}(s_{t+1}, \vect{a}') - Q(s_t,\vect{a}_t)\right],
\end{split}
\end{equation}

where $s_t$ is the state at time $t$, $\vect{a}_t$ is the action taken at time $t$, $r_{t+1}$ is the reward received at time $t+1$, $\alpha$ is the learning rate, and $\gamma$ is the discount factor that controls the importance of future rewards. The max operation selects the action with the highest Q-value in the next state. 
Deep Q-Learning uses a deep neural network $Q(s_t,\vect{a}; \vect{\theta})$ with parameters $\vect{\theta}$ to approximate the action-value function. The network takes a state $s$ as input and outputs Q-values for each action. The loss function for the network is defined as:

\begin{equation}
\begin{split}
    L(\theta) = \\
    \mathbb{E} \left[ (r_{t+1} + \gamma  \max_{\vect{a}'} Q^{target}(s_{t+1}, \vect{a}'; \vect{\theta}^-) - Q(s_t, \vect{a}_t, \vect{\theta}))^2\right]    
\end{split}
\end{equation}

where $Q_{\text{target}}$ is a target network with frozen parameters $\vect{\theta}^{-}$ that is used to generate the targets for the Q-values \cite{mnih_human-level_2015}. The Q-learning algorithm updates the parameters of the network by minimizing the loss function using stochastic gradient descent:

\begin{equation}
    \vect{\theta} \leftarrow \vect{\theta} - \alpha \frac{\partial L(\vect{\theta})}{\partial \vect{\theta}}
\end{equation}

As the agent interacts with the scenario, it will generate $(s^t,\vect{a}^t,r^t,s^{t+1})$ experiences that will be stored in a buffer memory that is fed to the optimization algorithms, adjusting the Q value to the new batch at each optimization step. For learning to be effective, an initial exploration phase of the state-action space is required. An $\epsilon$-greedy policy is used, in which each agent will take with a probability of $\epsilon$ a random action and with a probability $1 - \epsilon$ the optimal action indicated by the function $Q$, i.e., $\vect{a} = \max_{\vect{a}'} Q(s,\vect{a}'; \vect{\theta})$. To balance the exploration-exploitation of the network, $\epsilon$ is annealed from 1 (full random) to a minimum value of exploration $\epsilon_{min}$ (greedy).

\subsection{Observation function}


To define the observation $\vect{o}_j^t = \mathcal{O}(s^t)$ of an agent $j$, this work resorts to a visual description of the scenario. Each map of the scenario is discretized into an $[m,n]$-pixel matrix. Such discretization helps to reduce the complexity of the problem and, as mentioned in \cite{yanes_deep_2020}, it is convenient due to the spatial correlations between vehicle positions and areas of interest. Moreover, these visual states can be directly forwarded by convolutional neural policies as it will be explained later. Thus, the observation will be composed of 5 channel-images. 

\begin{enumerate}
    \item The mean of the Local GP $\vect{\hat \mu}(X)$. 
    \item The predictive uncertainty of the Local GP $\vect{\sigma}(X)$. 
    \item The navigation map $\mathcal{M}$ where values 1 indicate navigable positions.
    \item A null matrix with the position of vehicle $j$ with value 1.
    \item A null matrix with the positions of the other vehicles $j^-$ with value 1.
\end{enumerate}

All these images will be min-max normalized to be between 0 and 1.

\subsection{Reward function}

The definition of the reward function directly impacts the behavior of the fleet. Therefore, its definition is fundamental for acquiring the desired results \cite{sutton_reinforcement_2018}. The reward function $r(s,a)$, quantitatively determines how good or bad an action $a$ is in a state $s$ and must be aligned with the ultimate goal, which is to obtain a model as accurate as possible. 
To encourage agents to explore the environment, it is necessary to define a metric that evaluates the information gain from one instant to the next. In previous work, as in \cite{peralta_MULTIOBJECTIVE}, the utility function is based on the expected improvement of the function. This paper proposes to use two different reward functions and compare performance and alignment with the final goal.

First, a reward function similar to the one used in \cite{wildfires} for monitoring forest fire scenarios will be tested. In this work the reward is directly proportional to discovered ignited cells in a discretized wildfire scenario. Thus, this reward function takes into account the difference between the model at two consecutive time-steps. 
This paper will modify this function to reward the absolute value of the changes from a previous to a later mean of the model $\Delta \mu^t(\vect{X}) = \sum |\vect{\hat \mu}^t (\vect{X})- \vect{\hat \mu}^{t-1}(\vect{X})|$. This reward provides higher values when the changes between the posterior and the prior increases. This is motivated by the fact that obtaining data that changes its mean with respect to the prior means a better estimation, thus quantifying the quality of an action. The changes in the model is directly related to the Kullback-Leibler (KL) divergence as shown in \eqref{eq:KL} for a multivariate Gaussian distribution. 

A higher KL divergence will is directly related to the degree of change the GP model experience when adapting new data. 
Note that in \eqref{eq:KL} both the changes in uncertainties and means of two distributions impacts the divergence. 
Consequently, a increased divergence would imply that from two consecutive steps more valuable information is being utilized by the model.

\begin{equation}
\begin{aligned}
    KL(GP_1 \rightarrow GP_2) = \\
    \frac{1}{2}\left[\log \frac{\left|\Sigma_2\right|}{\left|\Sigma_1\right|}-d+\operatorname{tr}\left\{\Sigma_2^{-1} \Sigma_1\right\}+\left(\Delta \mu\right)^T \Sigma_2^{-1}\left(\Delta \mu \right)\right]
\end{aligned}
\label{eq:KL}
\end{equation}

In addition to this criterion, it is also proposed to use a reward function that benefit those actions that reduce the predictive uncertainty of the Gaussian process as much as possible. With the formulation proposed in Section \ref{section:gps}, the predictive uncertainty will be adjusted through the kernel parameters online by means of likelihood maximization. Where the lengthscale is smaller, reducing the uncertainty requires a larger number of samples and vice versa. Thus, the reward will be proportional to the absolute change in the uncertainty $\Delta \sigma^t(\vect{X}) = \sum |\vect{\hat \sigma}^t (\vect{X})- \vect{\hat \sigma}^{t-1}(\vect{X})|$. 
Note that the absolute value is taken because, as the hyperparameters are optimized, the covariance term changes, and the uncertainty could increase or decrease when new data are taken. Therefore, a change in the total uncertainty, regardless of the sign, also has an impact on the model improvement, since this shift in the new hyperparameters also contributes to the model accuracy.

\begin{figure}
    \centering
    \includegraphics[width=0.6\linewidth]{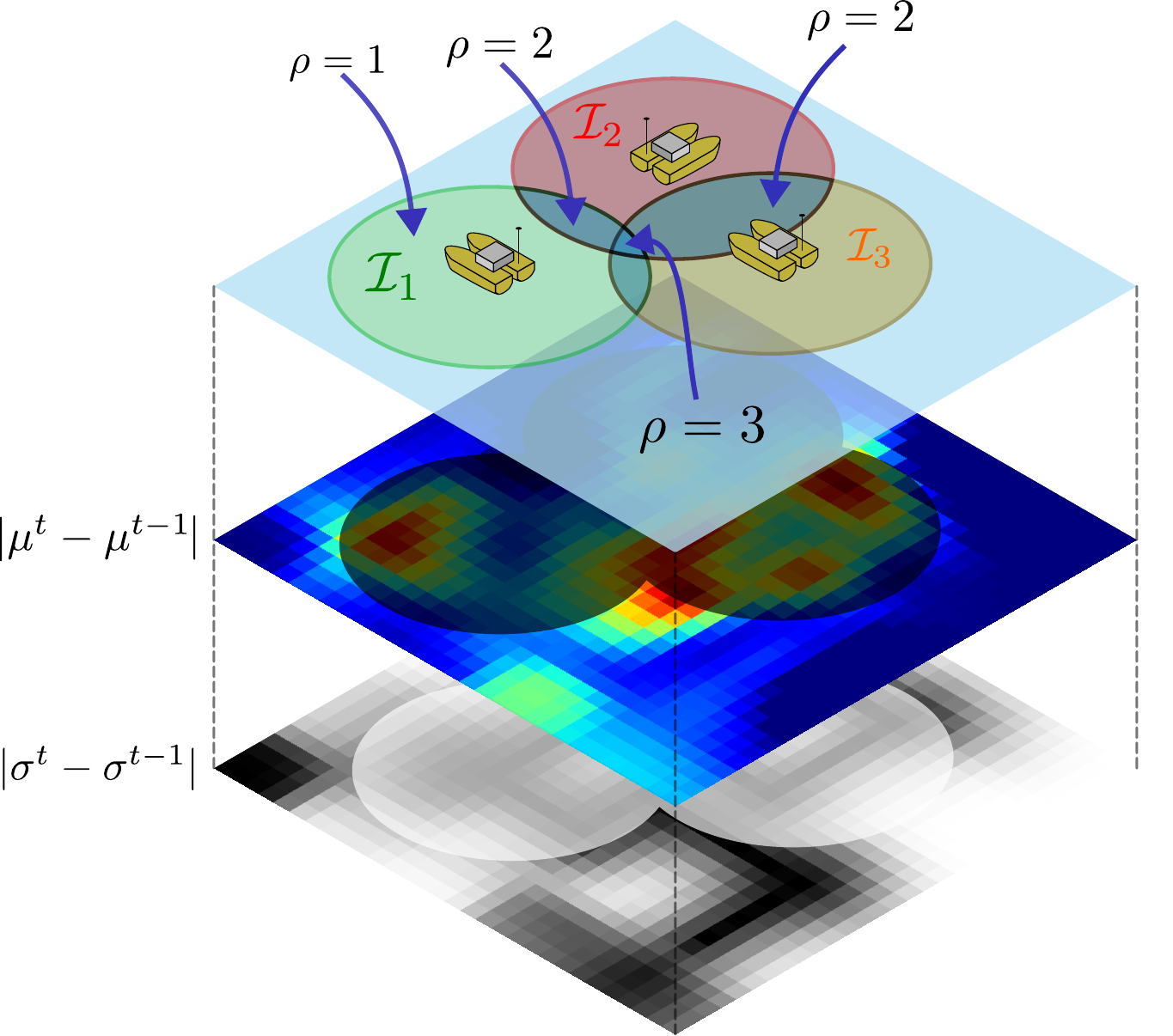}
    \caption{Influence areas $\mathcal{I}$ for every vehicle and its corresponding redundancy values $\rho$.}
    \label{fig:redundancy_zones}
\end{figure}

In the multi-agent case, it will be necessary to distribute the rewards according to the impact of each action on the total improvement of the model and to avoid that an agent who performs a bad action gets a biased reward for an action of another agent. Here, there are two important components.

First, it is a locality factor in the change. Specifically, it is only considered as part of the reward model those changes in an area of influence $\mathcal{I}$ of radius $R$ around each agent. 
This radius is the same as the level of influence of local GPs. Second, a redundancy factor $\rho(\vect{x})$ is introduced for each zone within the radius area of influence of the ASVs equal to the radius of the local GPs $\nu_k$. The value $\rho(\vect{x})$ is the number of agents who share the changes at that particular location $\vect{x}$. This redundancy factor will divide the aforementioned changes in the uncertainty or model, meaning that two ASVs that take the sample too close will receive half of the reward each. In Figure \ref{fig:redundancy_zones}, the reward parameters are visually depicted.

In summary, the two reward functions for every agent $j$ are:
\begin{equation}
    r_\mu(s^t,a_j^t) =  \sum_{\vect{x} \in \mathcal{I}_j} \left[ \frac{|\Delta \mu^t(\vect{x})|}{\rho_j(\vect{x})} \right]
    \label{eq:rew_mu}
\end{equation}

\begin{equation}
    r_\sigma(s^t,a_j^t) =   \sum_{\vect{x} \in \mathcal{I}_j} \left[ \frac{|\Delta \sigma^t(\vect{x})|}{\rho_j(\vect{x})} \right]
    \label{eq:rew_sigma}
\end{equation}

\subsection{Deep Safe Policy for multiagent training}

\begin{figure}[th]
         \centering
         \includegraphics[width=1\linewidth]{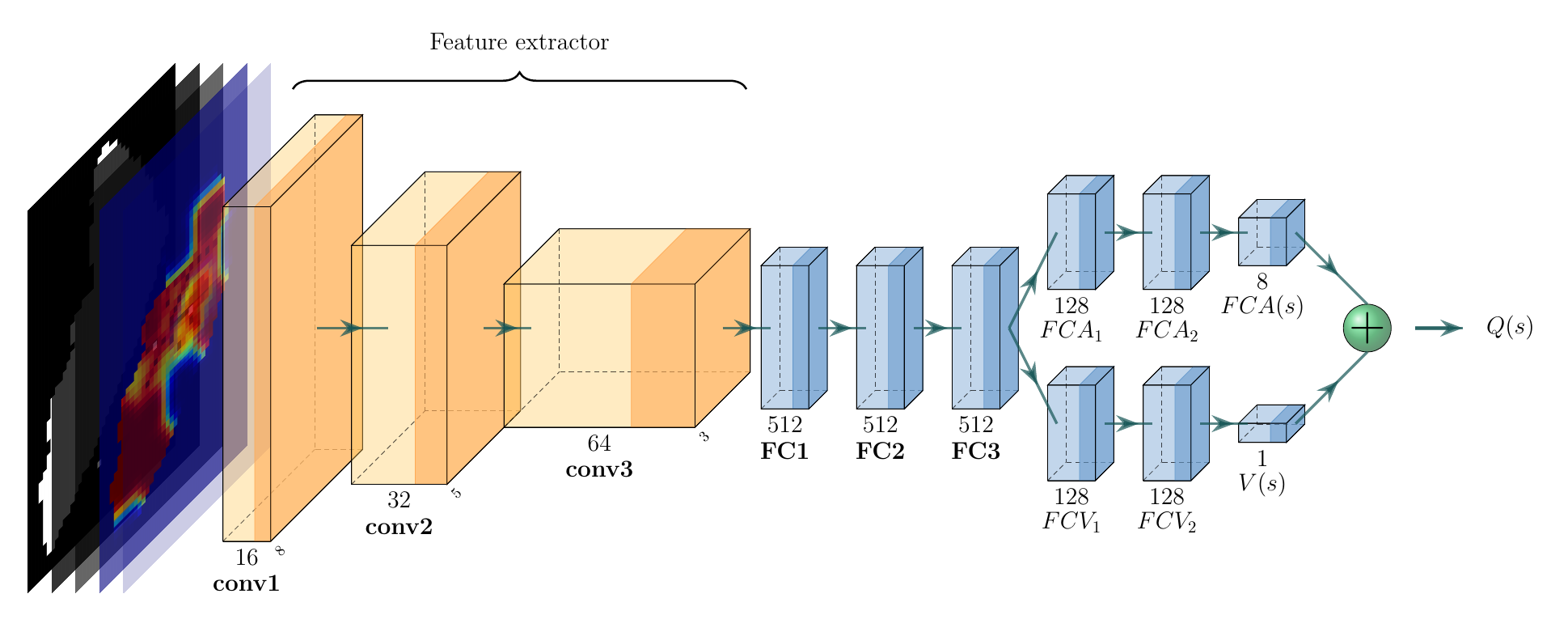}
         
        \caption{Dueling Neural Network architecture for the \textit{Q}-function representation. It is composed by an initial visual encoder and two heads: i) the Advantage head and the Value head. The outputs are the 8 \textit{Q}-values}
        \label{fig:network}
\end{figure}

\begin{figure}
    \centering
    \includegraphics[width=0.7\linewidth]{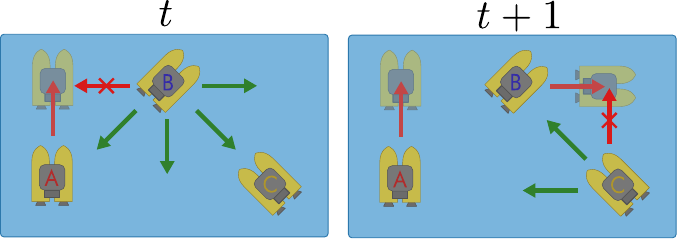}
    \caption{Consensus scheme for the safe action selection. At instant $t$, the agent with higher $Q$ chooses its action first. Then, the second agent take an action rejecting any that causes collision. This is repeated until all agents have decided the next action, and a consensus is reached.}
    \label{fig:consensus_scheme}
\end{figure}

For this multi-agent application, a single neural network for all agents will be used. This technique, called \textit{parameter sharing}, has been shown to be effective in previous work \cite{wildfires, yanes_multiagent_2021}. The difference of this proposal from previous work is that the same neural network is able to accommodate a different number of vehicles due to the egocentric formulation of the observation. As agents are interchangeable in perception and actions, all generated experiences are stored indistinctly in the experience buffer for later use in training. 

The proposed neural network is presented in Figure \ref{fig:network}. This neural network is composed of a first stage in the form of a convolutional encoder, as originally proposed in \cite{mnih_human-level_2015}. This stage extracts the visual features of each observation $o_j^t$ to produce an output $Q(o_j^t, \vect{a})$. The three consecutive convolutional layers are followed by three fully connected neural network layers. Following these, the values are unfolded into two heads for the computation of the value function $V(s)$ and the advantage function $A(s,\vect{a})$. This way of representing $Q(s,\vect{a})$ is based on \cite{wang_dueling_2015}, which allows for a better representation of the cumulative reward. Finally, according to \cite{wang_dueling_2015}, the value of $Q(s,\vect{a})$ is calculated as follows:

\begin{equation}
    Q(s,\vect{a}) = V(s) + \left( A(s,\vect{a}) - \frac{1}{|A(s,\vect{a})|} \sum_a A(s,\vect{a}) \right)
\end{equation}

While DRL effectively learns obstacle avoidance \cite{yanes_deep_2020}, deterministic computation can address actions leading to agent-scenario collisions \cite{censoring_yanes_2022}. However, collision between agents is more complex as simultaneous actions may cause conflicts. This work proposes a heuristic based on conditional decision-making to prevent such situations. Agents are sorted based on the highest joint value of $Q$, with the highest-Q agent taking an action without considering other agents. Subsequent agents consider the new position of the previous one, censoring $Q$ values leading to collisions with $-\infty$. Once actions are decided, movements are processed to prevent collisions (see Figure \ref{fig:consensus_scheme}). This heuristic relies on agent optimism to prioritize actions. In cases of random actions (following $\epsilon$-greedy policy), only safe actions are considered, avoiding collisions. Pseudo-code for the Safe Dueling DQL algorithm is provided in Algorithm \ref{alg:ddqn}, with the consensus subroutine outlined in Algorithm \ref{alg:safe_consensus}.

\begin{algorithm}
\caption{Safe Multiagent Double Deep Q-Learning Algorithm}
\label{alg:ddqn}
\begin{algorithmic}[1]
\State Initialize replay memory $D$ to capacity $|D|$
\State Initialize target Q-network $Q'$ with weights $\vect{\theta}' = \vect{\theta}$
\State Initialize policy network $Q$ with weights $\vect{\theta}$
\For{episode $=1$ to $E_{max}$}
    \State Reset environment
    \State Get initial observation $o_0 = \mathcal{O}(s_0)$
    \For{timestep $=1$ to $T$}
        \State $p \sim U(0,1)$
        \If{$p < \epsilon$}
            \State $\vect{a}_j \leftarrow SafeConsensus(U(0,1), \ldots, U(0,1))$
        \Else
            \State $\vect{a}_j \leftarrow SafeConsensus(Q(o_0,a), \ldots, Q(o_N,a))$
        \EndIf
        
        \State Execute action $\vect{a}_j$
        \State Observe rewards $r_j$ and new observations $o_j^{t+1}$
        \State Store every transition $(o_j^t,\vect{a}_j^t,r_j^t,o_j^{t+1})$ in $D$
        \State Sample random batch $B$ of $(o_j,\vect{a}_j,r_j,o_{j+1})$ from $D$
        \State Set $y_j = r_j + \gamma Q'(s_{j+1}, \arg\max_{\vect{a}} Q(s_{j+1},\vect{a};\vect{\theta});\vect{\theta}')$
        \State Update weights by minimizing the loss:
            \[
            \mathcal{L}(\vect{\theta}) = \frac{1}{B}\sum_{j=1}^{B} (y_j - Q(s_j,\vect{a}_j;\vect{\theta}))^2
            \]
        \State  $\vect{\theta}' \leftarrow \vect{\theta} \times \tau + (1 - \tau) \times \vect{\theta}'$ \Comment{Polyak target update}   
    \EndFor
    \State $\epsilon \leftarrow \min(\epsilon_{min}, \epsilon - d\epsilon)$.
\EndFor
\end{algorithmic}
\end{algorithm}

\begin{algorithm}
\caption{SafeConsensus algorithm}
\label{alg:safe_consensus}
\begin{algorithmic}[1]

\Require  Positions $P^t={\vect{p}^t_1,\vect{p}^t_2,\ldots,\vect{p}^t_N}$ of $N$ agents at time $t$
\Require Values $Q = \{Q_1,Q_2,\ldots,Q_N\}$ that weight each agent's action.
\State Initialize empty set of future positions $P^{t+1} := \emptyset$

\State Obtain order of agents' actions in decreasing order of their $Q$ values: $j_1,j_2,\ldots,j_N$, such that $\max Q_{j_1} \ge \max Q_{j_2} \ge \ldots \ge \max Q_{j_N}$.
\For{each agent $j$ in order of actions}

    \State Select greedy safe action
        \[
        \begin{aligned}
            \vect{a}_j = \arg\max_{\vect{a}\in A} Q_j(a) \\
            \text{subjected to:} \\
            \|(\vect{p}^t_j + \vect{a}_j) - \vect{p}'\|_2 \leq d_{safe} \quad \forall \vect{p}' \in P^{t+1} &
        \end{aligned}
        \]

    \State $P^{t+1} \leftarrow P^{t+1} \cup (\vect{p}^t_j + \vect{a}_j)$ \Comment{Update next fleet positions.}
    \State $A_{selected} \leftarrow A_{selected} \cup {\vect{a}_j}$ \Comment{Update consensus actions.}
\EndFor

\State \textbf{return} $A_{selected}$
\end{algorithmic}
\end{algorithm}

\section{Simulations and Results}

In this Section, all the experiments and simulations performed to validate the optimality of the algorithm are presented. First, the performance in terms of computation time and accuracy of Local GPs along with the classical approach of Global GPs is analyzed for both WQP Algae bloom benchmark. Then, the results of the DRL training are presented with both designed rewards and both benchmarks, with a discussion on the scalability of the proposal. Finally, the results are compared with other algorithms used in previous approaches in the literature.

All simulations and training were conducted on an Ubuntu 22.04 server, with 256Gb RAM, Dual Xeon CPU Scalable SP3 HPC, and two different GPUs: i) Nvidia RTX 3090 25GB, and Nvidia Quadro A4000 48GB. Python 3.10 and PyTorch were used for policy optimization. All simulation parameters and constraints are summarized in Table \ref{tab:sim_parameters}. The code will be available for reproduction of the results in \url{https://github.com/derpberk/}.

For the analysis of the results, the comparison between rewards and in three algorithms, the following metrics must be defined:

\begin{itemize}
    \item \textbf{Sum of Residuals (SoR)}: Defines the absolute sum of error between the estimated mean of the GP $\vect{\hat \mu}(\vect{x})$ and the ground truth value $f(\vect{x})$. This metric is intended to be minimized.
    \[
        SoR = \sum_{\vect{x} \in \vect{X}} | \vect{\mu}(\vect{x}) - f(\vect{x})|
    \]

    Different from other approaches \cite{peralta_MULTIOBJECTIVE, kathen_informative_2021}, the absolute error will be used to analyze the model accuracy, as the Mean Squared Error does not reflect well small improvements into the model error.

    \item \changes{\textbf{Normalized Sum of Residuals (nSoR)}: Normalization of the SoR respect to the amount of information of available in a Ground Truth:}
    \[
        \changes{nSoR = \frac{\sum_{\vect{x} \in \vect{X}} | \vect{\mu}(\vect{x}) - f(\vect{x})|}{\sum_{\vect{x} \in \vect{X}}{f(\vect{x})}}}
    \]
    
    \item \textbf{Average Error in $f(\vect{x}$ local maxima \footnote{The local maxima is computed in both ground truths using by applying a maximum filter with a neighborhood of $1.5~km$ and taking the locations where the magnitude of the second order derivative term is 0 using a Sobel filter.}}: Defines the mean error in the local maxima of the ground truth $f(\vect{x})$. This metric is useful to obtain a measurement of the error in the most biologically dangerous spots.

    \begin{equation}
    \begin{split}
        Avg. \; SoR \; f(\vect{x}^*) = \\
        \frac{1}{\text{\# peaks }\{f(\vect{X})\}} \sum_{\vect{x} \in \text{\#. peaks }\{f(X)\}} |\vect{\mu}(\vect{x}) - f(\vect{x}) |
    \end{split}
    \end{equation}

    \item \textbf{Max. Error in $f(\vect{x})$ local maxima}: Defines the max error in the local maxima of the ground truth $f(\vect{x})$. This metric provides a maximum bound of the error in the estimation of $f(\vect{x})$.
    \begin{equation}
    \begin{split}
        Max. \; SoR \; f(\vect{x}^*) = \\
        \max \{ | \vect{\mu(x)} - f(\vect{x}) | \; \forall x \in \text{\#. peaks }\{|f(\vect{X})|\} \} 
    \end{split}
    \end{equation}
    
\end{itemize}

\begin{table}[h]
    \centering
    
    \begin{tabularx}{\columnwidth}{X|X}
        \hline
         Parameter &  Value\\
         \hline
         Number of GPs $(K)$ & 18 \\
         Influence radius $(\nu_k)$ & $1.45~km$ \\
         $\ell$ interval $(\ell_{min}, \ell_{max})$ & (0.5, 10) \\
         Base uncertainty $(\sigma_0)$ & 1.0 \\
         Measurement noise $(\sigma_n)$ & $1\times10^{-5}$ \\
         Max. distance $(d_{max})$ & $29~km$ \\
         Safety distance $(d_{\mathrm{safety}})$ & $300 m$ \\
         Movement distance $(d_{meas})$ & $580 m$ \\
         Map size $(H,W)$ & $(58, 38)$ pixels\\
         
         \hline
    \end{tabularx}
    \caption{Environment and Model parameters}
    \label{tab:sim_parameters}
\end{table}

\subsection{Local Gaussian Process performance}

\begin{figure}[!ht]
    
         \centering
         \includegraphics[width=0.7\columnwidth]{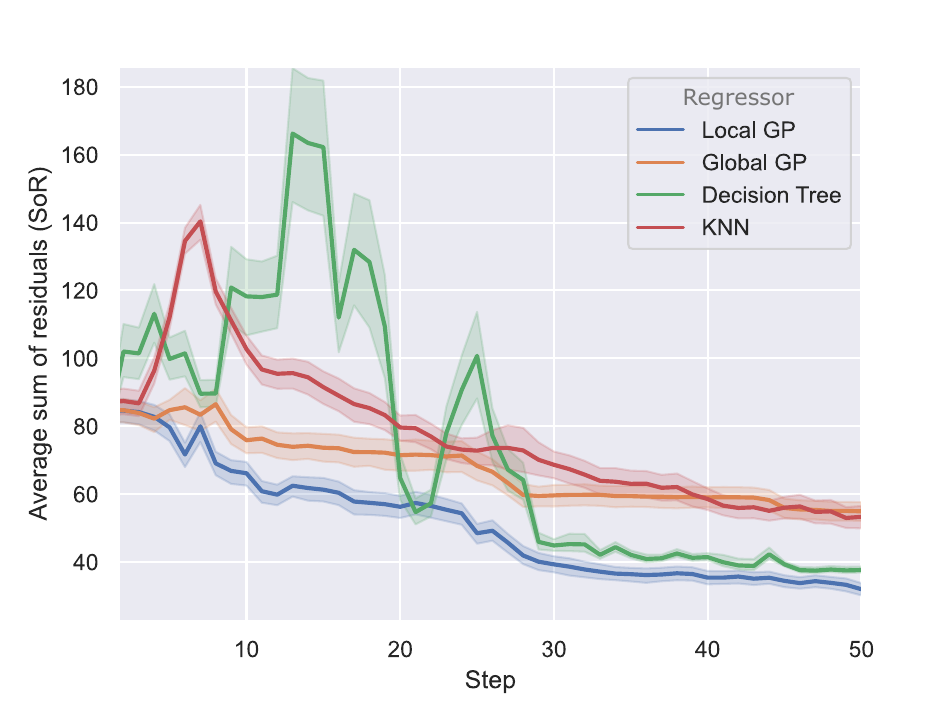}
         
         \caption{Accumulated inference time, between the proposed Local GP and the classic Global GP.}
         \label{fig:VSerror}

\end{figure}

\begin{figure}[!ht]
     \centering
     \includegraphics[width=\columnwidth]{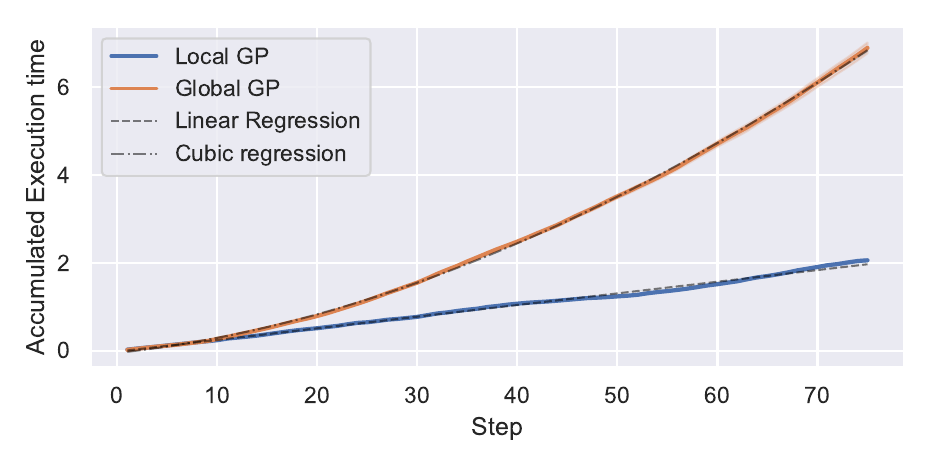}
     
    \caption{Comparison of error (a) between the proposed local GPs, and other ML algorithms, in the Ypacaraí scenario, for 50 different scenarios using different path planners. The colored area is the standard deviation.}
    \label{fig:VStime}
\end{figure}


Initially, 18 local GPs, distributed 2 km apart (Figure \ref{fig:Zones}), were validated with simulations. The radius of influence $\nu_k$ for each process was experimentally set at $1.45 km$ (5 pixels) based on the granularity of the scalar fields. In Figure \ref{fig:VSerror}, 50 missions were simulated for the Algae Bloom scenario with 3 agents using nonreactive path planners and various regression algorithms (Global GPs, kNN, Decision Tree) to assess local GPs with offline paths. Despite all algorithms collecting the same information at each instant, local GPs, on average, reduce the model estimation error by almost 20 points of SoR, a 33\% improvement from 40 samples. With an increasing number of samples, local GPs outperform a single global Gaussian process with the same information. Other regressors, like k-Nearest Neighbors (with $k = 5$), exhibit poor performance. Although Decision Tree accuracy drops with sufficient samples, the early inference process is slow to converge.


In Figure \ref{fig:VStime}, computation times are compared, revealing that a global Gaussian process experiences cubic growth with the number of samples. The depicted time represents the average cumulative time spent by the model server in model inference. This scalability issue intensifies with more agents and a larger sample size. Conversely, local GPs exhibit linear time growth with the number of samples. While this might not significantly impact small fleets, scalability becomes a concern as the number of samples and vehicles increases. It's important to note that, in this simulation, no concurrency in the optimization of local GPs was implemented. However, the fully parallelizable model optimization process could further reduce computation time.

\begin{figure}[ht]
    \centering
    \includegraphics[width=0.8\linewidth]{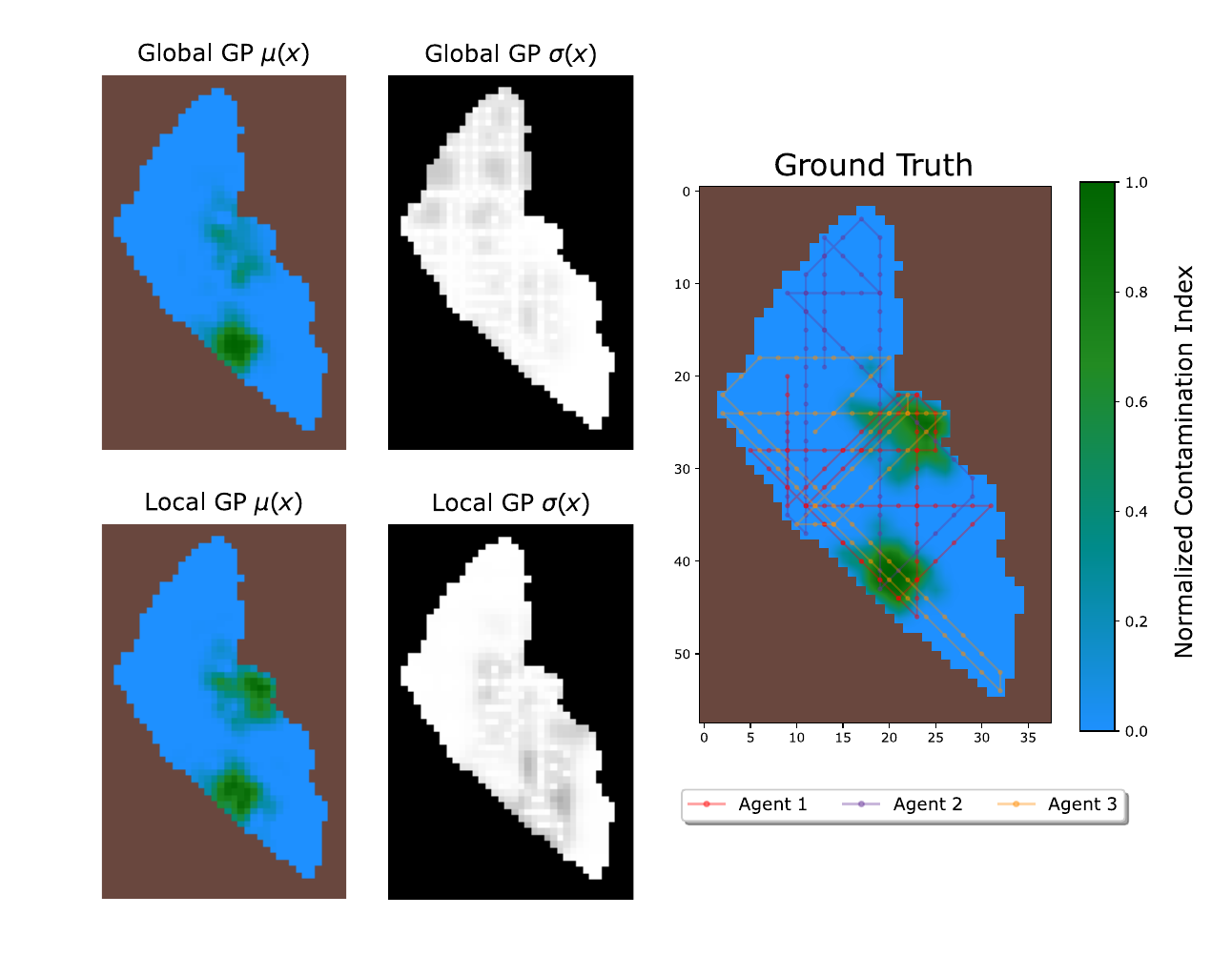}
    \caption{Comparison between the final model $\mu(X)$ and uncertainty $\sigma(X)$ using local GPs (down) and a global GP (up), with random explorative paths.}
    \label{fig:VSmaps}
\end{figure}

\begin{figure}[ht]
    \centering
    \includegraphics[width=1\linewidth]{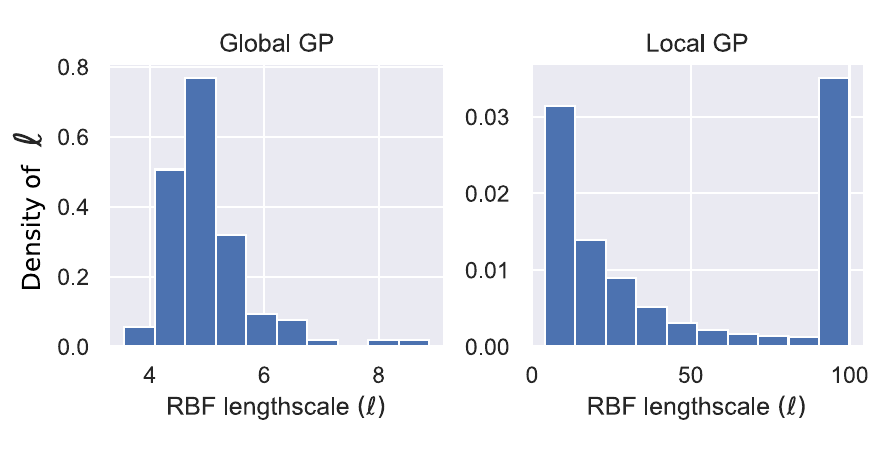}
    \caption{Histogram of RBF kernel length scales $\ell$ after likelihood optimization with 100 sample points using global GPs (left) and the proposed local GPs (right)}
    \label{fig:lengthscales}
\end{figure}

In Figure \ref{fig:VSmaps} it is presented an example of local vs. global Gaussian processes is presented giving the same information in the particular case of algae detection. It can be seen that Global GPs, as formulated in \cite{peralta_MULTIOBJECTIVE}, have estimation problems in areas of higher granularity. This occurs because the global GPs maximize the marginal likelihood of the kernel hyperparameters for the entire sample space. In the case of algae monitoring, a priori it seems to be that there are two distinct zones. Zones of low pollution concentration with high spatial correlation between samples (posterior $\ell$ is high) and zones of pollution hotspots with low correlation (posterior $\ell$ is small). As seen in Figure \ref{fig:lengthscales}, the global GP, converges in a small $\ell$ at the end of the mission, but as there are several highly correlated samples in low contaminated zones, the estimation in higher zones is affected. In the local GP case, the estimation in both zones is isolated due to the locality of the individual GPs.

By constraining a maximum $\ell$ to 100 (high enough to consider that all possible samples in the scenario are highly correlated), it is possible to obtain a distribution over the length scale $\ell$ for the conditions of the previous experiment. In Figure \ref{fig:lengthscales}, the resulting hyperparameter value of the posterior kernel can be observed after maximization of the likelihood at the end of every mission. In the Global GP, the maximum likelihood is found in values between 3 and 10. The less correlated samples will cause this length scale to drop to represent both low-correlated and high-correlated data. In the Local GP case, the histogram shows a multimodal distribution of the data. The parameter $\ell$ is found to maximize the marginal likelihood of the Local GPs with values lower than 10, and also with values in the limit of 100. This is translated into the environmental task into smaller values around algae blooms and higher length scales in zero-contaminated zones. This shows that the Local GP is able to bring up with richer representation of an arbitrary parameter distribution. Without loss of generality, it is reasonable to say that these GPs can be used to estimate more efficiently scalar maps with richer distribution of hyperparameters. 

\changes{In short, these local Gaussian processes generally alleviate the computational complexity with respect to the global counterpart. However, they still have $\mathcal{O}(N^3)$ complexity locally, and it may happen that an excessive number of samples in a single Gaussian process severely slows down the computation. On the other hand, in terms of the convergence of the processes, the convergence of each local process is still not guaranteed. This is closely related to the choice of the internal structure of the correlation represented by the Kernel. An inadequate selection of the kernel would certainly lead to a convergence of the local process. Although the locality of the proposed Gaussian processes limits the effect on global convergence, it is still a fundamental task to select the Kernel taking into account how the information we want to measure behaves.}

\subsection{DRL fleet training}

For DRL policy training, two reward functions (Section 4.4) were studied across fleets of 1 to 3 agents, each undergoing 10,000 missions of consistent duration (29 km - 50 steps). The evaluation covered different fleet sizes and benchmarks (WQP and Algae Bloom monitoring) to validate the approach under varied conditions. Hyperparameters were adopted from previous studies \cite{yanes_deep_2020} to streamline the training process. DDQL consistently converges to a similar policy with sufficient training episodes, falling within typical literature hyperparameter ranges. The $\epsilon$-greedy exploration policy maintains $\epsilon$ values from 1 to $\epsilon_{min}$ (0.05) in 50\% of the episodes to balance exploration and exploitation. Neural network training employs a batch size of 64 and a Learning Rate of 0.0001- In Table \ref{tab:learning_parameters} it is summarized all training parameters.

\begin{table}[]
    \centering
    
    \begin{tabularx}{\columnwidth}{X|X}
        \hline
         Parameter &  Value\\
         \hline
         Learning rate & $1\times10^{-4}$ \\
         \hline
         Batch size  & $64$ \\
         \hline
         $\epsilon_{min}$ & $0.05$ \\
         \hline
         $\epsilon$ episode anneal val. $d\epsilon$ & $1.9\times10^{-4}$ \\
         \hline
         $\tau$ & $1\times10^{-4}$ \\
         \hline
         Discount factor $\gamma$ & $0.99$ \\
         \hline
         Learning rate & $1\times10^{-4}$ \\
         \hline
         Activation Function & ReLU \\
         \hline
    \end{tabularx}
    \caption{Learning parameters for the DDQL algorithm.}
    \label{tab:learning_parameters}
\end{table}

  \begin{figure}[th]
    \centering
    \includegraphics[width=\linewidth]{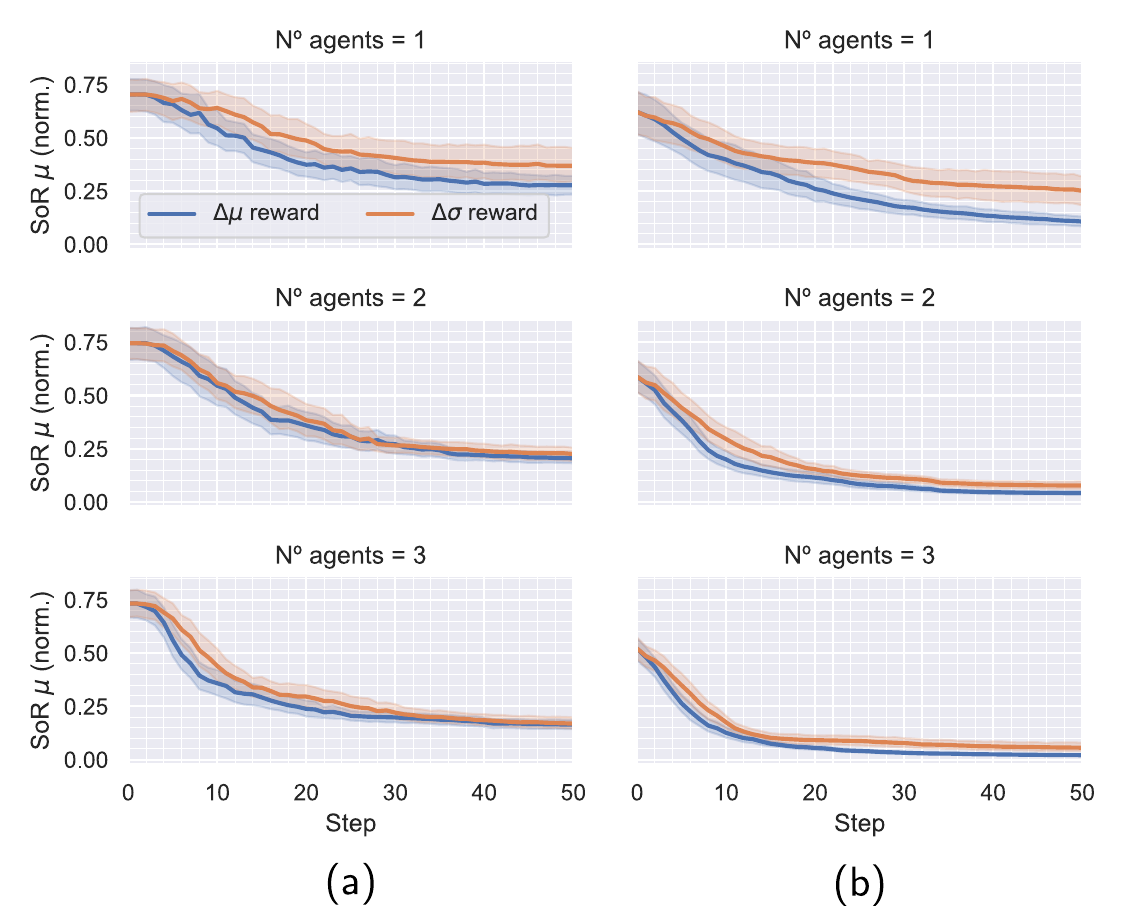}
    \caption{Estimation error (SoR) comparison between final policies trained with $\mu$-change reward (blue) and  $\sigma$-change reward (orange) in WQP (a) and Algae Bloom (b) benchmarks.}
    \label{fig:reward_comparison}
\end{figure}

\begin{figure}[th]
\centering
\includegraphics[width=\linewidth]{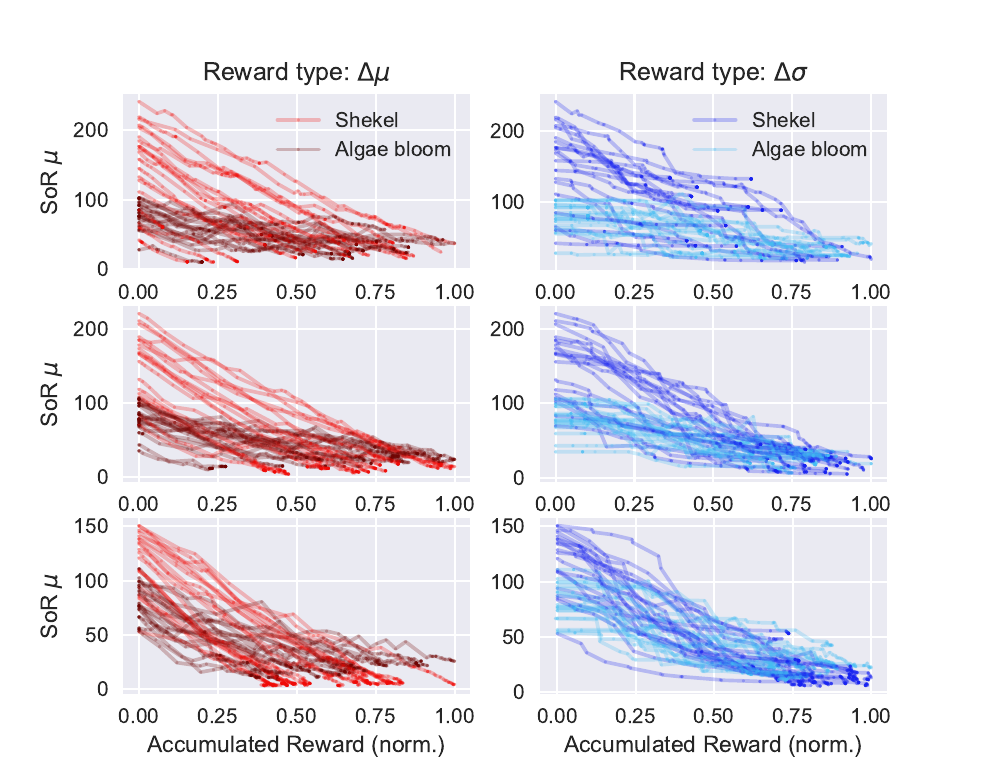}
\caption{Accumulated reward vs. the estimation error for every fleet size, reward type, and benchmark used.}
\label{fig:correlation}
\end{figure}

\begin{figure}[th]
\centering
\includegraphics[width=\linewidth]{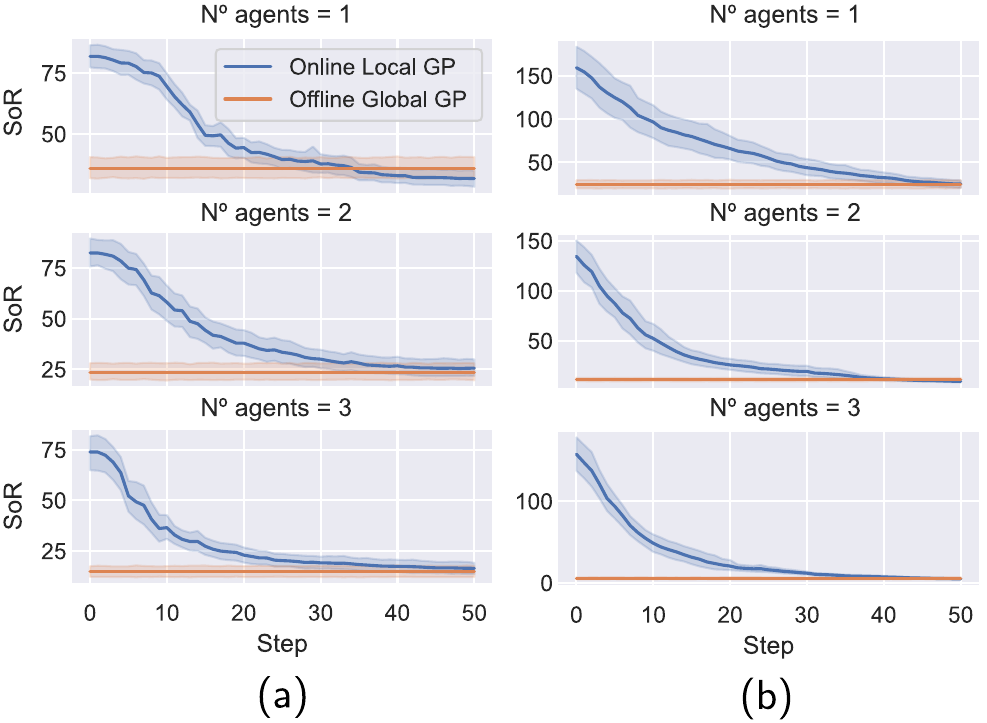}
\caption{Comparison of the online estimation error using the best $\mu$-change reward with local GPs, and the offline error using a global GP, with the same sample points, at the end of an episode in the WQP (a) and Algae Bloom (b) benchmarks.}
\label{fig:VSOffline}
\end{figure}


In Figure \ref{fig:reward_comparison}, both rewards described in Section 4.4 are compared for every possible combination of fleet size and benchmark, after 10.000 episodes of learning with the same parameters. In general, the reward relating to changes in $\vect{\hat \mu(X)}$ is more aligned with the objective of minimizing the estimation error. In the WQP case, an improvement of approximately 26\% with $p < 0.05$ using a Wilcoxon Ranked test can be observed. With 2 and 3 agents, the improvement is not significant, indicating that both rewards could provide similar performance. Nonetheless, the $\mu(X)$-changes reward obtains slightly better results earlier than the $\sigma(X)$-changes counterpart. The contrary happens in the Algae Bloom benchmark. The $\mu(X)$-change provides better and earlier performance in 3 fleet sizes. The error is $\sim 50\%$ better with all fleet sizes with $p < 0.05$. This makes sense if it is considered that in the $\mu(X)$-change reward, higher rewards are received only when the model update results in a significant change. As the model is prone to improve with every new sample, the net changes in the model are an accurate estimator of the error. In the $\sigma(X)$-change reward, agents receive higher rewards even when the prior and posterior models are close to each other. 


Figure \ref{fig:correlation} illustrates the cumulative reward for each fleet against prediction error, enabling the analysis of the correlation between reward and the goal of minimizing error using the R2 score. The $\mu(X)$-changes reward yields an R2 score of -4.54, while the $\sigma(X)$-reward results in an R2 score of -40.42. The $\mu(X)$-reward shows a more linear dependence with decreasing error, in contrast to the $\sigma(X)$-reward, which exhibits plateaus along each reward-error trajectory, indicating that exploring areas of low interest doesn't significantly improve the error. This comparison highlights the effectiveness of the $\mu(X)$-reward in error reduction. Table \ref{tab:uncertainty_final} displays the average predictive uncertainty values in the map. Encouraging a subtle reduction in predictive uncertainty leads to overconfidence in initially uninteresting areas. Additionally, the $\Delta \sigma$ reward results in a 24\% lower final average uncertainty compared to the $\Delta \mu$ reward.

\begin{table}[th]
\footnotesize
\centering
\begin{tabularx}{0.8\columnwidth}{|X|X|X|X|}
\hline
\textbf{GT} & \textbf{Reward} & \textbf{Nº Agents} & \textbf{Mean $\sigma(X)$} \\ \hline
    Algae & $\Delta \mu $ & 1 & 0.419 \\ \cline{3-4}
    & & 2 & 0.265 \\ \cline{3-4}
    & & 3 & 0.235 \\ \cline{2-4}
    & $\Delta \sigma$ & 1 & 0.338 \\ \cline{3-4}
    & & 2 & 0.147 \\ \cline{3-4}
    & & 3 & 0.104 \\ \hline
    WQP & $\Delta \mu$ & 1 & 0.252 \\ \cline{3-4}
    & & 2 & 0.152 \\ \cline{3-4}
    & & 3 & 0.024 \\ \cline{2-4}
    & $\Delta \sigma$ & 1 & 0.259 \\ \cline{3-4}
    & & 2 & 0.185 \\ \cline{3-4}
    & & 3 & 0.050 \\ \hline
\end{tabularx}
\caption{Average total uncertainty at the end of the missions, for the two reward types, fleet size, and benchmarks.}
\label{tab:uncertainty_final}
\end{table}


To validate DRL-trained policies and the use of local Gaussian processes, we compare online estimation errors with those obtained using a global Gaussian process with all collected data at the end (offline GP). Figure \ref{fig:VSOffline} depicts the SoR curves, showing that local GPs achieve results comparable to the global GP with full information. This indicates that, despite minor improvements, local GPs, being more flexible and time-efficient, can perform as well as global GPs with a proper policy. The synergy between DRL for sequential information gathering and local GPs as an efficient surrogate model generator is evident, as policies demonstrate low errors independently of the model used.

  \begin{figure}[htbp]
     \centering
     \includegraphics[width=0.75\columnwidth]{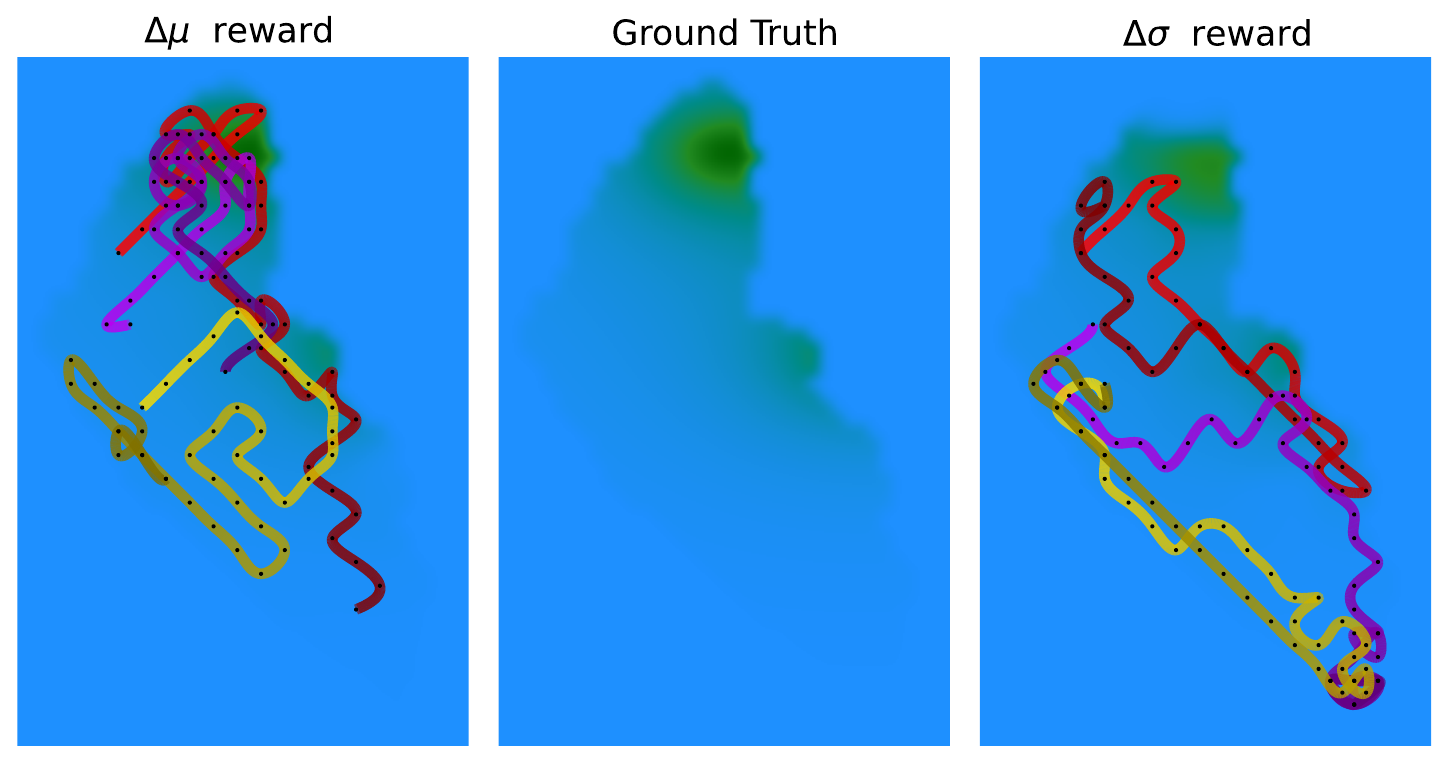}
     \caption{Resulting path of running a simulation with the $\Delta \mu$-reward (left) and the $\Delta \sigma$-reward for the same WQP monitoring scenario (middle).}
     \label{fig:pathsrewardshekel}
 \end{figure}

\subsection{Comparison with other algorithms}

 \begin{figure}[t]
     \centering
     \includegraphics[width=0.75\columnwidth]{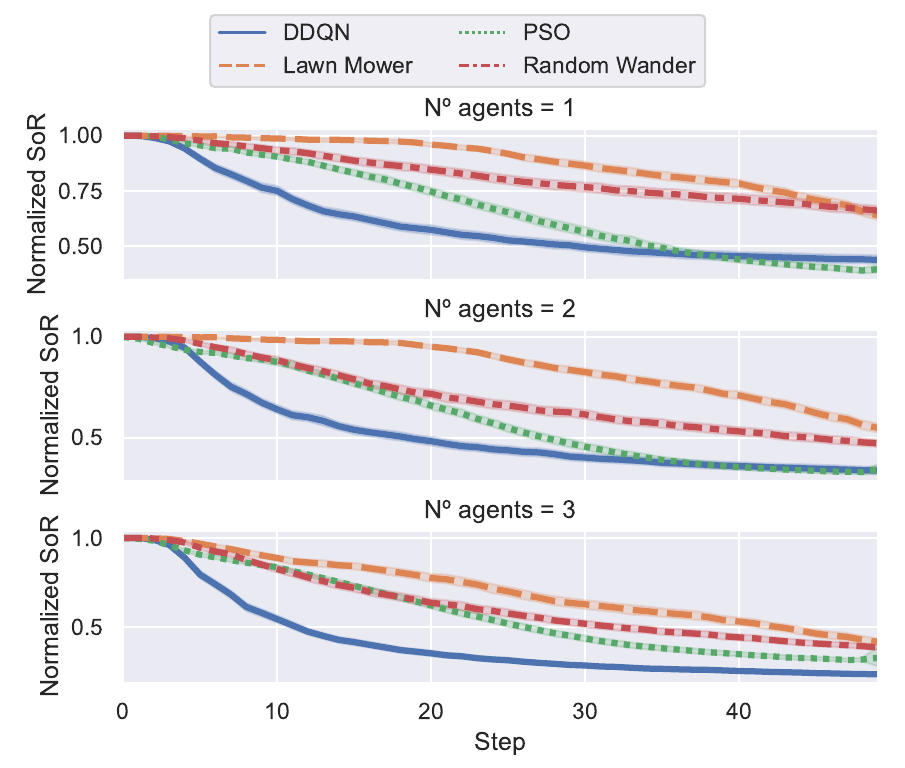}
     \caption{Estimation error between other algorithms (LMPP, RWPP, PSO) and our DDQL trained policies with the $\mu$-change reward, for the WQP monitoring benchmark.}
     \label{fig:ErrorShekel}
 \end{figure}

 \begin{figure}[t]
     \centering
     \includegraphics[width=0.75\columnwidth]{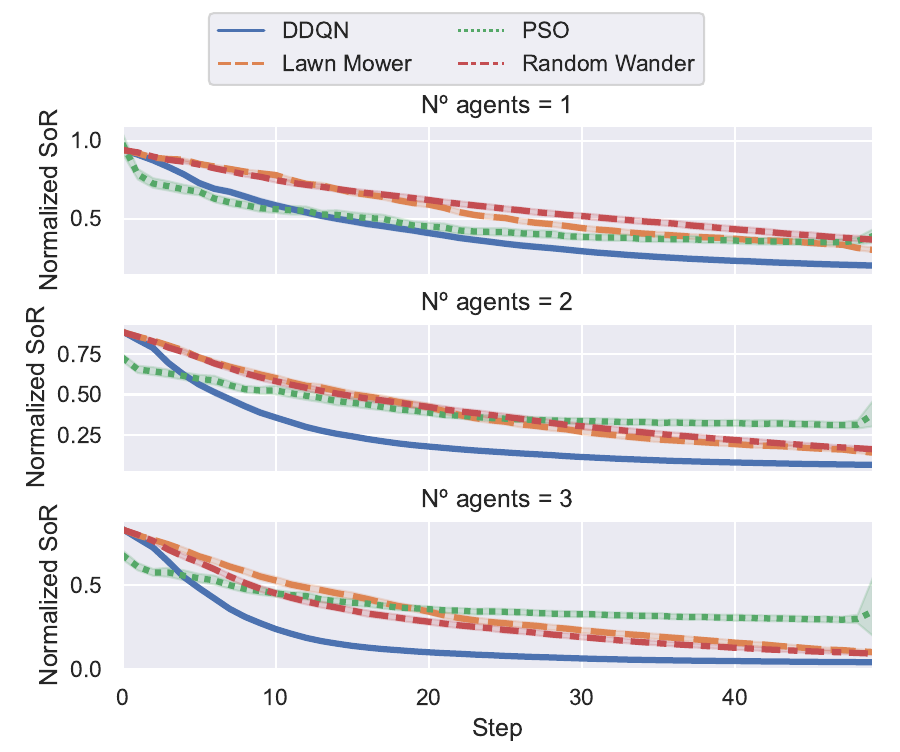}
     \caption{Estimation error between other algorithms (LMPP, RWPP, PSO) and our DDQL trained policies with the $\mu$-change reward, for the Algae Bloom monitoring benchmark.}
     \label{fig:ErrorAlgae}
 \end{figure}

This work also compared the trained policies with other path planning algorithms to validate the results. For this comparison, the $\Delta \mu$ reward trained policies is used, as they show the best performance in terms of exploration and intensification of high-interest zones. The comparison is made with 3 different path planners:

\begin{enumerate}
    \item \textbf{Lawn Mower Path Planner (LMPP):}  The LMPP consists of maximizing the coverage of vehicles by taking samples in parallel lines. Every agent select a random initial direction to initialize the path. When an obstacle is reached, the agent travels back in the reverse direction in a parallel line. This algorithm will use Local GPs as a model for the contamination.
    \item \textbf{Random Wanderer Path Planner (RWPP)}: This approach generates random exploratory paths by selecting a direction of exploration. Every agent select a random free-obstacle direction until a new obstacle is met. Then, the agent selects a direction different from the previous direction to avoid retracing its steps. This algorithm also uses Local GPs as a model for the contamination.
    \item \textbf{GP-Enhanced Particle Swarm Optimization (EG-PSO)}: This approach is taken directly from \cite{kathen_informative_2021}. In this approach, every vehicle is a particle that will change its speed proportional to 4 distances: i) the distance to the maximum uncertainty, ii) the distance to the maximum sampled value observed by the agent, iii) the distance to the maximum global value samples by the fleet, and iv) the distance to the maximum value predicted by the model.  
\end{enumerate}

\changes{Up to 300 different simulation were conducted for every Ground Truth type and with every benchmark. We used 6 different seeds to reduce the effect of epistemic uncertainty in the results. For a fair comparison, this evaluation set of ground truths will be different from any other episode experimented during training for the DQL.} In Figures \ref{fig:ErrorShekel} and \ref{fig:ErrorAlgae}, the online estimation error is represented for the WQP and Algae Bloom benchmarks, respectively. It is observed that, in general, the DRL is able to obtain better results. In Tables \ref{tab:WQP_comparison_results} (WQP benchmark) and \ref{tab:algae_comparison_results} (Algae Bloom benchmark), the metrics for the aforementioned simulations are presented, with the mean and std. deviation values of each algorithm.

\begin{figure}[htbp]
     \centering
     \includegraphics[width=\columnwidth]{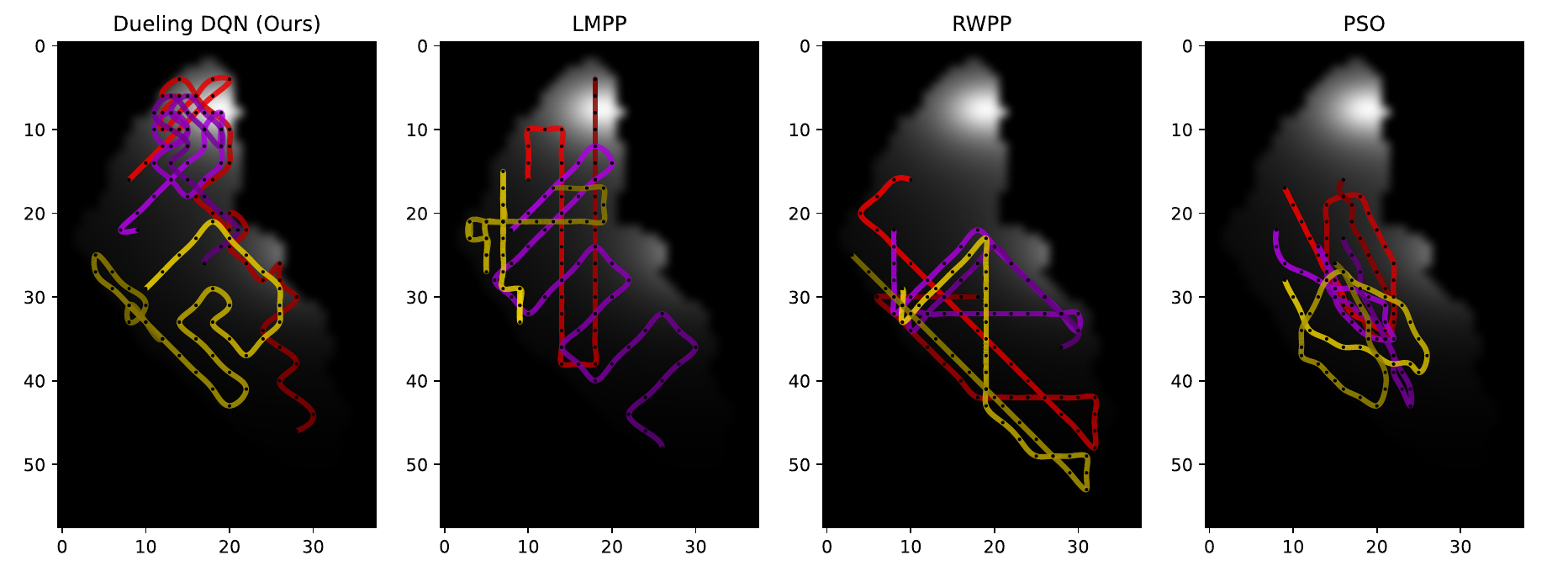}
     \caption{Resulting paths for all other algorithms (LMPP, RWPP, PSO) and our DDQL trained policy with the $\mu$-change reward, for the WQP monitoring benchmark.}
     \label{fig:pathsshekel}
 \end{figure}

\begin{figure}[htbp]
     \centering
     \includegraphics[width=\columnwidth]{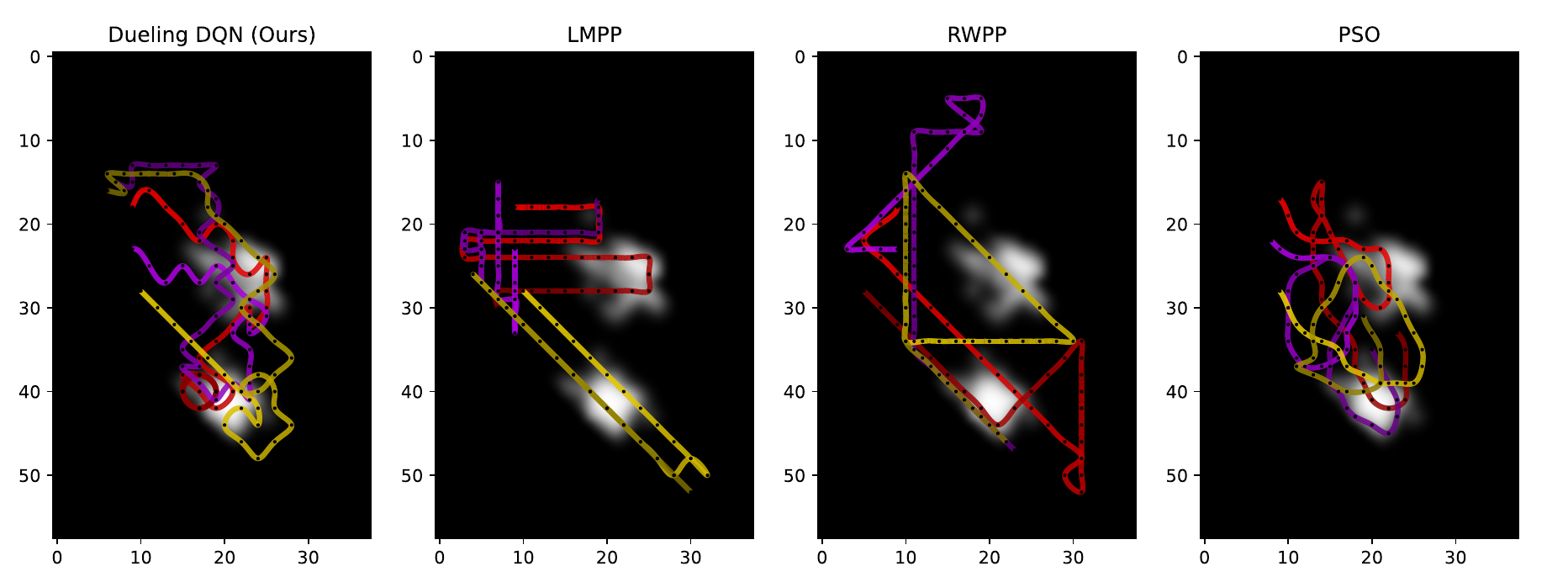}
     \caption{Resulting paths for all other algorithms (LMPP, RWPP, PSO) and our DDQL trained policy with the $\mu$-change reward, for the Algae Bloom monitoring benchmark.}
     \label{fig:pathsalgae}
 \end{figure}

\begin{table*}[h]
    \centering
    \caption{Metric comparison between algorithms for the WQP modeling mission. The highlighted metrics refer to the best performance algorithm}
    \resizebox{0.8\linewidth}{!}{
        \begin{tabular}{|c|c|c c|c c|c c|c c|c c|}
            \hline
            \multirow{2}{*}{Algorithm} & \multirow{2}{*}{$N_{agents}$} & \multicolumn{2}{c|}{$SoR (33\%)$} & \multicolumn{2}{c|}{$SoR (66\%)$} & \multicolumn{2}{c|}{$SoR (100\%)$} & \multicolumn{2}{c|}{$\text{Avg. SoR in } f(x^*)$} & \multicolumn{2}{c|}{$\text{Max. SoR in } f(x^*)$} \\
            \cline{3-12}
            & & Mean & Std. Dev. & Mean & Std. Dev. & Mean & Std. Dev. & Mean & Std. Dev. & Mean & Std. Dev. \\
            \hline\hline
            \multirow{3}{*}{D-DQL} 
& 1 & \textbf{0.62} & 0.18 & \textbf{0.48} & 0.17 & 0.44 & 0.17 & \textbf{0.20} & 0.17 & \textbf{0.35} & 0.30\\ \cline{2-12}
& 2 & \textbf{0.53} & 0.18 & \textbf{0.39} & 0.18 & \textbf{0.34} & 0.16 & \textbf{0.15} & 0.14 & \textbf{0.24} & 0.29\\ \cline{2-12}
& 3 & \textbf{0.40} & 0.15 & \textbf{0.27} & 0.09 & \textbf{0.24} & 0.07 & \textbf{0.09} & 0.08 & \textbf{0.17} & 0.16\\ \cline{2-12}
            \hline\hline
            
            \multirow{3}{*}{PSO \cite{kathen_informative_2021}} 
& 1 & 0.81 & 0.14 & 0.52 & 0.19 & \textbf{0.40} & 0.08 & 0.37 & 0.26 & 0.39 & 0.30\\ \cline{2-12}
& 2 & 0.75 & 0.15 & 0.41 & 0.15 & 0.34 & 0.06 & 0.22 & 0.14 & 0.24 & 0.18\\ \cline{2-12}
& 3 & 0.71 & 0.13 & 0.40 & 0.14 & 0.33 & 0.10 & 0.21 & 0.15 & 0.39 & 0.28\\ \cline{2-12}
            \hline\hline
            
            \multirow{3}{*}{LMPP} 
& 1 & 0.98 & 0.07 & 0.84 & 0.16 & 0.63 & 0.19 & 0.43 & 0.25 & 0.67 & 0.32\\ \cline{2-12}
& 2 & 0.97 & 0.08 & 0.79 & 0.17 & 0.54 & 0.20 & 0.35 & 0.25 & 0.58 & 0.35\\ \cline{2-12}
& 3 & 0.83 & 0.18 & 0.60 & 0.22 & 0.41 & 0.17 & 0.24 & 0.20 & 0.40 & 0.31\\ \cline{2-12}
            \hline\hline
            
            \multirow{3}{*}{RWPP} 
& 1 & 0.88 & 0.16 & 0.75 & 0.20 & 0.66 & 0.20 & 0.46 & 0.26 & 0.69 & 0.32\\ \cline{2-12}
& 2 & 0.77 & 0.19 & 0.58 & 0.19 & 0.47 & 0.17 & 0.28 & 0.21 & 0.46 & 0.31\\ \cline{2-12}
& 3 & 0.70 & 0.20 & 0.49 & 0.17 & 0.39 & 0.13 & 0.21 & 0.15 & 0.39 & 0.28\\ \cline{2-12}
            \hline
            
        \end{tabular}
        \renewcommand{\arraystretch}{1}
    }
    \label{tab:WQP_comparison_results}
    \end{table*}

\begin{table*}[htbp]
    \centering
    \caption{Metric comparison between algorithms for the Algae Bloom modeling mission. The highlighted metrics refer to the best performance algorithm}
    \resizebox{0.8\linewidth}{!}{
        \begin{tabular}{|c|c|c c|c c|c c|c c|c c|}
            \hline
            \multirow{2}{*}{Algorithm} & \multirow{2}{*}{$N_{agents}$} & \multicolumn{2}{c|}{$SoR (33\%)$} & \multicolumn{2}{c|}{$SoR (66\%)$} & \multicolumn{2}{c|}{$SoR (100\%)$} & \multicolumn{2}{c|}{$\text{Avg. SoR in } f(x^*)$} & \multicolumn{2}{c|}{$\text{Max. SoR in } f(x^*)$} \\
            \cline{3-12}
            & & Mean & Std. Dev. & Mean & Std. Dev. & Mean & Std. Dev. & Mean & Std. Dev. & Mean & Std. Dev. \\
            \hline\hline
            \multirow{3}{*}{D-DQL} 
& 1 & \textbf{0.47} & 0.15 &          0.27 & 0.13 & \textbf{0.20} & 0.14 & \textbf{0.15} & 0.16 & \textbf{0.25} & 0.27\\ \cline{2-12}
& 2 & \textbf{0.23} & 0.09 & \textbf{0.10} & 0.05 & \textbf{0.07} & 0.04 & \textbf{0.05} & 0.06 & \textbf{0.09} & 0.12\\ \cline{2-12}
& 3 & \textbf{0.13} & 0.06 & \textbf{0.06} & 0.03 & \textbf{0.04} & 0.02 & \textbf{0.04} & 0.04 & \textbf{0.06} & 0.07\\ \cline{2-12}
            \hline\hline
            
            \multirow{3}{*}{PSO \cite{kathen_informative_2021}} 
& 1 & 0.50 & 0.23 & 0.38 & \textbf{0.18} & 0.38 & 0.22 & 0.41 & 0.24 & 0.53 & 0.26\\ \cline{2-12}
& 2 & 0.44 & 0.25 & 0.33 & 0.16  & 0.38 & 0.22 & 0.43 & 0.22 & 0.60 & 0.23\\ \cline{2-12}
& 3 & 0.39 & 0.18 & 0.32 & 0.15 & 0.35 & 0.29 & 0.26 & 0.21 & 0.41 & 0.35\\ \cline{2-12}
            \hline\hline
            
            \multirow{3}{*}{LMPP} 
& 1 & 0.65 & 0.16 & 0.41 & 0.15 & 0.30 & 0.13 & 0.32 & 0.27 & 0.48 & 0.35\\ \cline{2-12}
& 2 & 0.49 & 0.18 & 0.24 & 0.15 & 0.14 & 0.10 & 0.18 & 0.20 & 0.29 & 0.28\\ \cline{2-12}
& 3 & 0.42 & 0.17 & 0.21 & 0.13 & 0.10 & 0.08 & 0.12 & 0.17 & 0.21 & 0.26\\ \cline{2-12}
            \hline\hline
            
            \multirow{3}{*}{RWPP} 
& 1 & 0.66 & 0.15 & 0.49 & 0.17 & 0.36 & 0.16 & 0.37 & 0.27 & 0.54 & 0.34\\ \cline{2-12}
& 2 & 0.47 & 0.17 & 0.28 & 0.16 & 0.16 & 0.11 & 0.20 & 0.22 & 0.30 & 0.29\\ \cline{2-12}
& 3 & 0.33 & 0.16 & 0.17 & 0.13 & 0.09 & 0.09 & 0.12 & 0.18 & 0.19 & 0.24\\ \cline{2-12}
            \hline
            
        \end{tabular}
    }
    \label{tab:algae_comparison_results}
    \end{table*}


In the WQP task, notable improvement, particularly in the multi-agent case, is observed. Offline algorithms like LMPP or RWPP consistently reduce errors over time. DQL demonstrates adaptability, prioritizing actions with higher short- and long-term rewards, resulting in a 45\% average improvement over other algorithms. LMPP, while robust with sufficient distance, tends to make inefficient movements in the WQP benchmark due to a lack of trajectory changes in low-interest areas. Random exploration behaves similarly to LMPP, but RWPP, being more exploratory, changes the monitoring front more frequently. These findings affirm that an effective IPP enhances overall modeling accuracy, even though using local Gaussian processes provides an advantage. It's reasonable to consider a significant dependence on the modeling method and information richness acquired.

\begin{figure}[htbp]
    \centering
    \includegraphics[width=\linewidth]{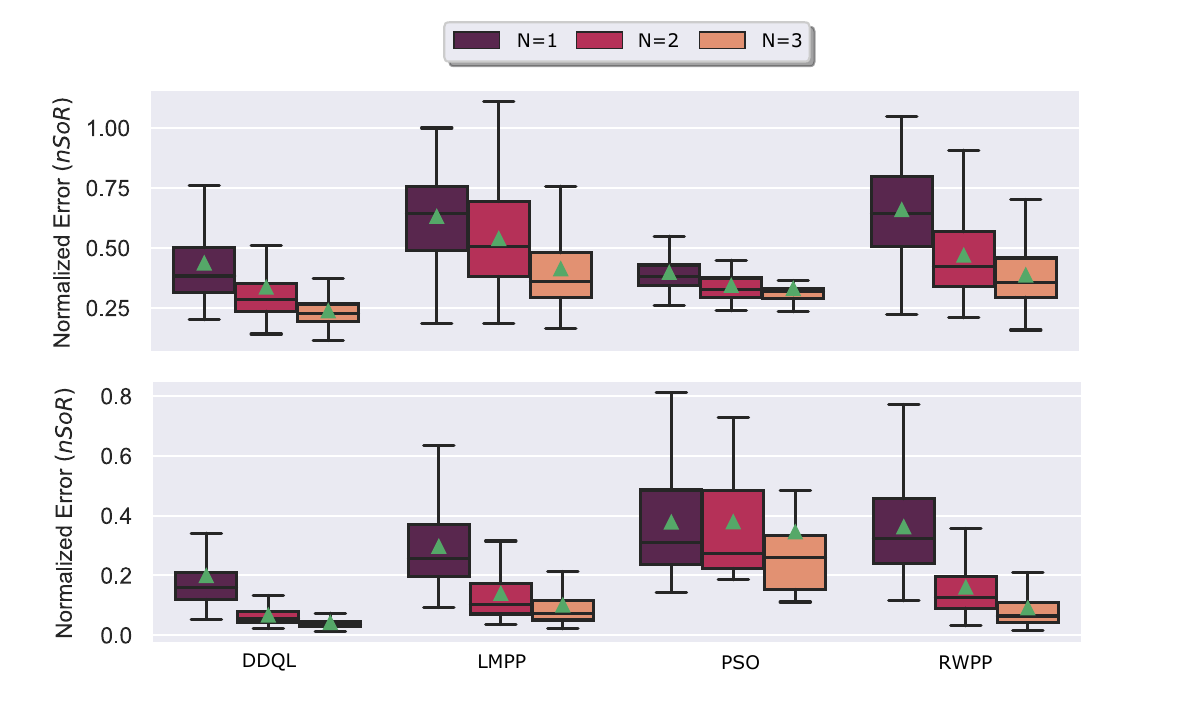}
    \caption{Box-plot representation of the final normalized error for 300 experiments with different algorithms. The upper plot corresponds with WQP benchmark. The lower plot corresponds with the Algae Bloom benchmark.}
    \label{fig:boxplot}
\end{figure}

\changes{In the PSO model discussed in \cite{kathen_informative_2021}, a significant challenge arises in partitioning the search space among agents. Deployed in close proximity within zones $Z_1, Z_2, Z_3$ (as shown in Figure \ref{fig:gts}), agents exhibit a gradient-descent behavior in the multi-agent case. Despite resembling a single agent due to similar local gradients caused by the absence of a dispersion mechanism, PSO achieves good convergence, especially in the single agent scenario and at episode completion (refer to Figure \ref{fig:boxplot}). Nevertheless, our proposed algorithm, leveraging the adaptive capacity of the DDQL policy, demonstrates faster convergence even if it doesn't surpass PSO in the single-agent case (see Figure \ref{fig:ErrorShekel}, Table \ref{tab:WQP_comparison_results}).}


The proposed algorithm presents good properties with respect to the time of exploration and mean error with respect to the other algorithms. This translates into an improvement of the error with the second-best algorithm (EG-PSO), on average for every fleet size, of a (32\%, 15\%, 6\%) at 33\%, 66\% and 100\% respectively, of the distance traveled. In the particular case of $N = 3$, with 43\% of the path distance budget traveled (16 steps of 48 samples), the estimation error is 27.75\% better than the second-best algorithm at that point (PSO). Regarding the metrics related to the error in the maximums of the benchmark function ($\text{Avg. SoR in } f(x^*)$ and $\text{Max. SoR in } f(x^*)$), it can be seen that the proposed algorithm is able to reduce the average error in the contamination maxima better than the other algorithms. It has been observed a 27\% better estimation in those points on average with the DQL with respect to the second best algorithm (RWPP).


In the second benchmark, with algae monitoring, the results also indicate the advantage of using DRL. This second benchmark is more difficult to monitor, and this is reflected in the improved results of DRL over the other algorithms. In terms of estimation at the end of a mission, the results show a 17\% improvement over the second-best algorithm (LMPP) on average for all fleet sizes. In this new case, the offline algorithms present a robust but in most cases inefficient result. It can be observed in Figure \ref{fig:ErrorAlgae} that during the course of a mission, both present a similar estimation error (no significant difference over a Wilcoxon ranking test. From this it can be inferred that, in this benchmark, the paths have to be less explorative and more exploitative in the search for algae sources. \changes{This compounds the need for a more comprehensive route planner that prioritizes high-interest areas in pursuit of a better model. In the end, the synergy between local Gaussian processes and an intelligent planner stands out when information is sparse in the search space, by the GPs reach convergence earlier. In other words, to obtain a good model with less samples and less movements.}

The PSO algorithm, on the other hand, due to its high dependence on the local gradients of each agent and the fact that the global maximum uncertainty point is insufficient to guide the fleet, is unable to perform on this benchmark as conceived in \cite{kathen_informative_2021}. The paths result in the absence of local gradients in a purely random scan unable to find the algae centroids in many cases. When it comes to estimation, PSO utilizes a global Gaussian process. However, the initial set of highly correlated samples causes the global lengthscale to quickly reach its upper limit with the first few samples. This, in turn, makes it difficult for the model to converge later on, especially in the presence of new samples.

\changes{The DRL algorithm, in the Algae Bloom benchmark, shows better improvement compared to previous approaches (see Table \ref{tab:algae_comparison_results}). With respect to the second-best result (LMPP), it is obtained an average improvement of a (49\%, 55\%, 48\% ) at 33\%, 66\% and 100\% respectively, of the distance traveled, among all fleet sizes. Improvement is also translated into higher speeds of model convergence. With every agent included in the fleet, the DRL is able to reduce the error earlier (35\% faster on average). In the estimation of pollution maximums, the DRL finds the maximums with higher average precision (up to 42\% lower error at these points for 3 vehicles) and with a lower maximum error (up to 40\% lower error at these points for 3 vehicles). This indicates that, in this new benchmark, the performance is robust and provides a good estimate of the errors in the most contaminated areas. This will be convenient when an early warning system requires to track dangerous spots of algae blooms for prevention and bath restrictions.}


\section{Conclusion}

This paper presents a framework for training and deploying multi-agent fleets of autonomous surface vehicles for environmental monitoring missions. The framework combines local Gaussian processes for model estimation and deep policies trained with DRL for decision-making. Two stochastic benchmark simulators were introduced to validate results for different environmental monitoring missions.

Local Gaussian processes significantly improve model computation time and yield a 30\% average reduction in estimation error with various path planners. These local models excel in estimating scalar fields of varying smoothness and multimodal hyperparameter distributions, demonstrating effectiveness in challenging scenarios like algae bloom monitoring. Combining different local models enhances granularity in estimating scalar functions with distinct local properties, especially beneficial in scenarios with steep gradients, such as algae monitoring.

Deep policies, derived from a Deep Reinforcement Learning algorithm, along with a consensus decision method, yield efficient monitoring policies complying with safety constraints during training. The proposed consensus mechanism is scalable, allowing for the independent adjustment of the number of agents and observations. Studying an appropriate reward function, based on the total net change of the model $\Delta \mu$, improves training efficiency by 26\% and achieves a 27\% average enhancement in benchmarks with other path planning algorithms. Specialization of the DRL algorithm in each mission results in an additional 30\% reduction in errors with improved efficiency and measurement redundancy. The combination of Gaussian processes with DRL emerges as a superior strategy for this mission type, with the reward function supporting online fleet retraining under real conditions based on the model's convergence estimation rather than the real ground truth.

\section{Acknowledgments}

We want to specially thank Thomas Wiedemann and all the Swarm Exploration Team of the Communications and Navigation Institute of the DLR in Munich for the advice and fruitful discussions about this research.

This work has been partially funded by the FPU-2020 PhD Grant (Formación del Profesorado Universitario) of Samuel Yanes Luis. This work was also possible thanks to the research stay of Samuel Yanes Luis in the Institute of Navigation and Communications of the German Aerospace Center (Weißling, Germany) from Sep. to Dec. 2022 under the supervision of Dmitriy Shutin. 


\bibliographystyle{IEEEtran}
\bibliography{cas-refs}
\end{document}